\documentclass[preprint,12pt]{elsarticle}
\biboptions{numbers,sort&compress}
\usepackage{a4wide}
\usepackage{latexsym,amssymb,amsmath,epsfig,amsfonts,algorithm}
\usepackage[title]{appendix}
\usepackage{eucal}
\usepackage{multirow,multicol,array,bm}
\usepackage{pgf-pie}
\usepackage{mathtools}
\usepackage{tabularx, booktabs, makecell, caption}
\usepackage{siunitx}
\usepackage[colorlinks=true,allcolors=blue]{hyperref}
\usepackage{float}
\biboptions{square,comma,numbers}
\floatstyle{plaintop} 
\restylefloat{table} 
\usepackage{caption}
\usepackage[utf8]{inputenc}
\usepackage[english]{babel}
\usepackage[noend]{algpseudocode}
\usepackage{tikz}
\usetikzlibrary{arrows}
\usepackage{bbm}
\usepackage{graphicx}
\usepackage{algorithm}
\usepackage{algpseudocode}
\usepackage[skip=0.333\baselineskip]{subcaption}
\usepackage{hyperref}
\usepackage{amsmath}
\usepackage{algorithm}
\usepackage{siunitx}
\usepackage{comment}
\usepackage{soul}
\usepackage{forest}
\useforestlibrary{edges}
\usetikzlibrary{arrows.meta, positioning}
\usepackage{amssymb}
\usepackage{amsmath}
\usepackage{amsthm}
\usepackage{lineno}

\begin{document}
\makeatletter
\def\ps@pprintTitle{}
\makeatother

\begin{frontmatter}

\title{Forecasting India’s Demographic Transition Under Fertility Policy Scenarios Using hybrid LSTM–PINN Model} 

\author{
Subarna Khanra\textsuperscript{1}, 
Vijay Kumar Kukreja\textsuperscript{1}, 
Indu Bala\textsuperscript{2,*} \\
\textsuperscript{1}Sant Longowal Institute of Engineering and Technology, Longowal, Punjab 148106, India \\
\textsuperscript{2}School of Computer and Mathematical Sciences, The University of Adelaide, Adelaide SA 5005, Australia \\
\textsuperscript{*}Corresponding author: indu.bala@adelaide.edu.au
}

\begin{abstract}
Demographic forecasting remains a fundamental challenge for policy planning in rapidly evolving nations such as India, where fertility transitions, policy interventions, and age-structured dynamics interact in complex ways. In this study, we present a hybrid modelling framework that integrates policy-aware fertility functions into a Physics-Informed Neural Network (PINN) enhanced with Long Short-Term Memory (LSTM) networks to capture physical constraints and temporal dependencies in population dynamics. The model is applied to India’s age-structured population from 2024 to 2054 under three fertility-policy scenarios: continuation of current fertility decline, stricter population control, and relaxed fertility promotion. The governing transport–reaction partial differential equation is formulated with India-specific demographic indicators, including age-specific fertility and mortality rates. PINNs embed the core population equation and policy-driven fertility changes, while LSTM layers improve long-term forecasting across decades. Results show that fertility policies substantially shape future age distribution, dependency ratios, and workforce size. Stricter controls intensify ageing and reduce labour force participation, whereas relaxed policies support workforce growth but increase population pressure. Our findings suggest that the hybrid LSTM–PINN is an effective approach for demographic forecasting, offering accuracy with interpretability. Beyond methodological novelty, this work provides actionable insights for India’s demographic policy debates, highlighting the need for balanced fertility interventions to ensure sustainable socio-economic development.
\end{abstract}

\end{frontmatter}

\section{Introduction}
\label{sec1}

Forecasting demographic evolution has emerged as a central challenge for nations undergoing rapid socio-economic transformation. Declining fertility rates and gradually increasing life expectancy reshape national age structures, with profound implications for labour markets, social security systems, and long-term economic sustainability \cite{asao2024japan}. In India, the world's most populous nation, the interplay between fertility transitions, population ageing, and shifting policy priorities creates a highly dynamic and difficult-to-predict demographic landscape \cite{chen2023changing}. Understanding and anticipating these demographic shifts is critical for evidence-based policymaking, particularly as fertility policy interventions increasingly influence population trajectories \cite{bongaarts2022fertility}.
Traditional approaches to population modelling have relied on classical compartmental frameworks \cite{bernardi2025heterogeneously,diekmann2023systematic,zhao2023comparative} and PDE-based transport models \cite{losanova2023boundary}. These methods provide strong theoretical foundations grounded in demographic theory and offer interpretable mathematical structures. However, they often struggle to capture complex, nonlinear dynamics and lack the flexibility to adapt to rapidly changing demographic patterns observed in contemporary developing economies.

The emergence of machine learning techniques has introduced new possibilities for demographic forecasting. Recent studies have applied data-driven approaches to population dynamics \cite{chen2025data,halder2023numerical,halder2024higher}, demonstrating improved predictive accuracy in certain contexts. However, these purely empirical models lack interpretability and frequently disregard fundamental demographic principles, limiting their reliability for long-term policy-sensitive forecasting. This trade-off between flexibility and theoretical consistency remains a persistent challenge in the field.

Physics-Informed Neural Networks (PINNs) \cite{raissi2019physics,karniadakis2021physics,tao2025analytical} have emerged as a promising approach that bridges this gap by embedding governing equations directly into deep learning architectures. This ensures that model predictions remain consistent with underlying physical or demographic laws while retaining the flexibility of neural networks. When integrated with Long Short-Term Memory (LSTM) networks \cite{hochreiter1997long,gers2002learning}, which excel at capturing temporal dependencies, hybrid LSTM-PINN frameworks can model systems where historical patterns strongly influence future trajectories.

The effectiveness of LSTM-PINN  models has been demonstrated across diverse domains. Cho et al. \cite{cho2022lstm} applied this framework to lithium-ion battery temperature estimation, showing enhanced robustness under limited data conditions. Kim et al. \cite{kim2025performance} demonstrated improved seismic response prediction by integrating temporal learning with physics-informed constraints. Wang et al. \cite{wang5101380hybrid} achieved more accurate fatigue life predictions in materials science by combining temporal pattern recognition with physical laws. These applications collectively demonstrate that LSTM-PINN can effectively balance data-driven flexibility with theoretical consistency. Jiang et al. \cite{jiang2025neighbour} proposed the LSTM-PINN Model for a physically consistent prediction of Soil Moisture and Water Retention Curves. Zhang et al. \cite{zhang2025temperature} stated compensation Method for Fiber-Optic Gyroscope Based on LSTM-PINN.

In demographic research, several studies have examined India's fertility dynamics and their determinants. Gandotra \cite{gandotra1998fertility} provided foundational insights into fertility patterns and reproductive behaviour across socioeconomic groups. Mahapatra et al. \cite{mahapatra2020migration} explored how migration influences family planning decisions in high-fertility regions. Dharmalingam et al. \cite{dharmalingam2014determinants} analysed drivers of fertility decline, emphasizing the roles of education and socio-economic development. These studies establish that India's demographic evolution is shaped by complex interactions between policy interventions, socioeconomic factors, and cultural dynamics.

Despite these advances, significant gaps remain in the demographic forecasting literature. First, while LSTM-PINN have proven effective in engineering and physical sciences, their application to demographic forecasting—particularly age-structured population dynamics—remains largely unexplored. Second, existing demographic models for India either rely on classical compartmental approaches that lack flexibility or employ purely data-driven methods that sacrifice theoretical consistency and interpretability. Third, no comprehensive framework currently exists that integrates age-structured population dynamics with explicit modelling of temporal policy influences, such as fertility interventions, within a unified physics-informed deep learning architecture \cite{bernardi2025heterogeneously}.

This gap is particularly critical for India, where rapid demographic transitions are reshaping dependency ratios, workforce composition, and social security demands. Policymakers require forecasting tools that not only predict future population structures but also quantify how alternative policy scenarios would alter demographic trajectories. Such tools must be both theoretically grounded and flexible enough to capture the nonlinear, path-dependent nature of demographic change.

This study addresses these gaps by developing a novel LSTM-PINN framework specifically designed for age-structured population forecasting in India. Our approach integrates demographic theory through physics-informed constraints while leveraging LSTM networks to capture long-term temporal dependencies inherent in fertility and mortality transitions. The framework explicitly incorporates policy interventions as temporal variables, enabling scenario-based forecasting under alternative fertility policies.

The specific objectives of this study are:

\begin{itemize}
    \item \textbf{Framework Development:} Develop and implement an LSTM-PINN architecture for age-structured population dynamics that embeds demographic conservation laws and fertility-mortality relationships within the neural network structure.
    
    \item \textbf{Model Calibration and Forecasting:} Calibrate the model using India's demographic data spanning 2024 to 2054, incorporating historical fertility and mortality patterns, and generate scenario-based forecasts under alternative fertility policy interventions.
    
    \item \textbf{Policy-Relevant Insights:} Provide interpretable outputs that quantify how fertility policy choices influence age structure evolution, dependency ratios, and workforce composition, offering actionable guidance for Indian policymakers and a transferable methodological framework for other developing countries facing similar demographic transitions.
\end{itemize}

By combining the theoretical consistency of physics-informed learning with the temporal modelling capabilities of LSTM networks, this study contributes a methodologically novel and practically relevant approach to demographic forecasting. The framework's emphasis on interpretability and policy sensitivity distinguishes it from existing methods and addresses the critical need for evidence-based demographic planning tools in rapidly transforming economies.
\paragraph{Novelty} As far as the available literature indicates, this is the first study to develop an age-structured demographic forecasting framework specific to India that jointly combines a transport–reaction PDE with both PINN and LSTM–PINN architectures. Unlike existing demographic applications that are either country-agnostic or focused on China, our model embeds policy-aware fertility functions, India’s recent NFHS/SRS-based demographic indicators, and multiple prospective policy scenarios (baseline, voluntary two-child, and enhanced family planning). Furthermore, we complement the hybrid modelling with population pyramids and LIME-based explanations, providing an interpretable, policy-relevant tool for long-term demographic planning in India and other developing countries.

\section{Mathematical Formulation of Age-Structured Population Dynamics}
\label{sec 2}

To capture the interaction between age-specific biological processes, long-term temporal dependencies, and India’s evolving fertility policy environment, we formulate the demographic system using an age-structured transport--reaction partial differential equation (PDE). This formulation enables explicit representation of births, deaths, and ageing while providing a natural interface for policy-aware extensions and neural network embedding. The PDE framework offers three key advantages: (i) interpretability rooted in demographic theory, (ii) compliance with conservation laws, and (iii) seamless integration with physics-informed neural networks (PINNs) for constrained learning \cite{gandotra1998fertility,mahapatra2020migration,suriyakala2016factors}.

India-specific demographic components—including age-specific fertility patterns, mortality schedules, and heterogeneous policy responses—are introduced directly within this mathematical structure. Empirical distributions and behavioural determinants affecting fertility and mortality \cite{subaiya2011demographics,james2011india} are incorporated alongside documented policy interventions such as family-planning incentives, awareness campaigns, and state-level restrictions \cite{rouyer1987political,aspalter2002population,diamond2010population,jindal2018mid}. This creates a coherent and policy-sensitive foundation for long-horizon scenario forecasting.

A central modelling innovation is the integration of policy-aware fertility functions and LSTM-driven temporal adjustments directly into the governing PDE. This ensures that fertility and mortality evolve not only with age and time but are dynamically influenced by past demographic behaviour and historical policy trajectories. The resulting PDE structure forms the analytical kernel of the hybrid LSTM–PINN model.

\subsection{Policy-Aware Fertility Function Embedding} 
A key innovation of our framework lies in the explicit embedding of fertility policy effects into the age-structured fertility  . 
Classical demographic models typically assume fertility rates $f(a,t)$ depend only on biological age $a$ and calendar time $t$, independent of policy interventions \cite{tao2025lstm}. 
However, this assumption overlooks how fertility incentives, restrictions, and socio-economic shocks reshape reproductive behaviour dynamically. 
To capture this, we introduce a policy-aware fertility function:
\begin{equation}
f(a,t) = f_{0}(a,t) + \Delta f(a,t; \mathcal{P}(t)),
\end{equation}
where
\begin{itemize}
    \item $f_{0}(a,t)$: baseline fertility rate, estimated from demographic data in the absence of policy,
    \item $\mathcal{P}(t)$: policy signal function encoding interventions (e.g., incentives, penalties, two-child norms, awareness campaigns),
    \item $\Delta f(a,t; \mathcal{P}(t))$: policy-dependent perturbation term that modifies fertility trajectories.
\end{itemize}

\subsubsection*{Baseline Fertility Profile}
The baseline fertility is expressed as:
\begin{equation}
f_{0}(a,t) = \beta(t)\, g(a),
\end{equation}
where
\begin{equation}
g(a) = \frac{1}{\sigma \sqrt{2\pi}} \exp\!\left(-\frac{(a-\mu)^2}{2\sigma^2}\right), \quad a \in [15,49],
\end{equation}
models the biological fertility window with peak fertility age $\mu$ and dispersion $\sigma$. 
Here, $\beta(t)$ captures long-term socio-economic trends in fertility decline or increase, independent of policy.

\subsubsection*{Policy Signal Function}
We embed time-varying policy interventions into a continuous function \cite{cho2022lstm}:
\begin{equation}
\mathcal{P}(t) = \sum_{k=1}^{K} \alpha_k \, h_k(t-t_k),
\end{equation}
where
\begin{itemize}
    \item $t_k$: introduction year of policy $k$,
    \item $\alpha_k$: signed strength (positive for promotion, negative for restriction),
    \item $h_k(\cdot)$: temporal response kernel (e.g., exponential decay, logistic adoption curve).
\end{itemize}
For example, a fertility promotion incentive can be modelled as
\begin{equation}
h_k(t-t_k) = \frac{1}{1+\exp(-\gamma (t-t_k))},
\end{equation}
with adoption rate $\gamma$.

\subsubsection*{Policy-Induced Perturbation}
We couple $\mathcal{P}(t)$ with age-specific sensitivity:
\begin{equation}
\Delta f(a,t; \mathcal{P}(t)) = \theta(a)\, \mathcal{P}(t),
\end{equation}
where $\theta(a)$ is an age-weighting function, such as a spline or piecewise polynomial that amplifies policy impact in target age groups (e.g., 20--30 years) \cite{chakravorty2021family}.  
Thus, the final fertility function becomes
\begin{equation}
f(a,t) = \beta(t)\, g(a) \;+\; \theta(a)\, \mathcal{P}(t).
\end{equation}

\subsubsection*{ Embedding Fertility into the Boundary Condition}
The fertility function modifies the boundary condition of the transport--reaction PDE governing population dynamics \cite{karniadakis2021physics, aspalter2002population}:
\begin{equation}
n(0,t) = \int_{a_{\min}}^{a_{\max}} f(a,t)\, n(a,t)\, da,
\end{equation}
where $n(a,t)$ is the population density at age $a$ and time $t$, and $n(0,t)$ denotes the number of newborns at time $t$. 
Here, fertility is explicitly policy-dependent, ensuring demographic dynamics reflect both biological constraints and government interventions.

\subsection{Transport–Reaction PDE with LSTM-Enhanced Temporal Dependencies}

The evolution of an age-structured population can be described by a transport--reaction partial differential equation (precisely known as the McKendrick–von Foerster PDE), where transport accounts for ageing and reaction terms capture fertility and mortality processes \cite{keyfitz1997mckendrick, mico2006age}. For India-specific demographics, the governing PDE is expressed as

\begin{equation}
\frac{\partial n(a,t)}{\partial t} + \frac{\partial n(a,t)}{\partial a} = -\mu(a,t)\, n(a,t) + B(a,t),
\label{eq:pde_basic}
\end{equation}

where $n(a,t)$ denotes the population density of age $a$ at time $t$, $\mu(a,t)$ is the age-specific mortality rate, and $B(a,t)$ represents the inflow of newborns derived from fertility contributions.  

Traditional formulations assume that $\mu(a,t)$ and $B(a,t)$ are deterministic functions of time and age. However, demographic evolution in India is significantly influenced by \textit{long-range temporal dependencies}, such as gradual policy impacts, generational fertility shifts, and historical mortality trends. These cannot be adequately captured by PDEs alone, which are inherently local in time \cite{dilao2006weak}. 
To address this gap, we propose an LSTM-Based Temporal Operator $\mathcal{L}_{\text{LSTM}}$ that captures the latent temporal dynamics of both fertility and mortality functions discussed in the next subsection.

\subsubsection{LSTM-Based Temporal Operator}
To capture long-term memory effects, we introduce LSTM-driven temporal refinements:

\begin{equation}
\mu(a,t) = \mu_{\text{base}}(a,t) + \mathcal{L}_{\text{LSTM}}^{(\mu)}(h_{t-1}, x_t),
\label{eq:mu_lstm}
\end{equation}

\begin{equation}
B(a,t) = B_{\text{policy}}(a,t) + \mathcal{L}_{\text{LSTM}}^{(B)}(h_{t-1}, x_t),
\label{eq:B_lstm}
\end{equation}

where $\mu_{\text{base}}(a,t)$ and $B_{\text{policy}}(a,t)$ are policy-aware baseline functions (as defined in Subsection~2.1), $x_t$ represents observed demographic covariates (e.g., fertility indicators, policy shifts, socio-economic signals),$h_{t-1}$ is the hidden state propagated through the LSTM across historical time steps, and $\mathcal{L}_\text{LSTM}$ is learned temporal operator.  

By embedding $\mathcal{L}_{\text{LSTM}}$ into the PDE system, the model simultaneously respects the mechanistic transport--reaction structure and leverages LSTM memory for capturing \textit{multi-decade temporal influences}. The resulting hybridized formulation is:

\begin{equation}
\frac{\partial n(a,t)}{\partial t} + \frac{\partial n(a,t)}{\partial a} =
- \Big[ \mu_{\text{base}}(a,t) + \mathcal{L}_{\text{LSTM}}^{(\mu)} \Big] n(a,t) 
+ \Big[ B_{\text{policy}}(a,t) + \mathcal{L}_{\text{LSTM}}^{(B)} \Big].
\label{eq:hybrid_pde}
\end{equation}

This formulation provides three major advantages:
\begin{enumerate}
    \item \textbf{Physical fidelity}: The transport--reaction PDE preserves interpretability and consistency with demographic theory.  
    \item \textbf{Policy adaptivity}: Fertility and mortality are dynamically adjusted to reflect the evolving political landscape of India.  
    \item \textbf{Temporal expressiveness}: LSTM embedding ensures that historical population patterns shape long-term forecasts.  
\end{enumerate}

Therefore, unlike conventional PDE-based demographic models where fertility remains static, our formulation dynamically embeds fertility with policy signals through $f(a,t)$. This dynamic coupling enables simulation of alternative fertility policy scenarios in India and forecasting of their long-term cascading effects on age-structured population evolution. By embedding $f(a,t)$ within the proposed hybrid LSTM-PINN, policy interventions are not treated as external inputs but rather as integral components that actively shape the PDE-constrained dynamics learned by the model.

This hybrid PDE formulation serves as the analytical backbone for the LSTM-PINN architecture introduced in Section~\ref{sec 3}. The architecture operates through a synergistic integration: the PINN component enforces Equation~\ref{eq:hybrid_pde} by minimizing physics-based residuals, ensuring adherence to demographic conservation laws, while the LSTM component captures history-dependent patterns to refine fertility and mortality predictions. This dual-mechanism design ensures that demographic trajectories simultaneously respect both the mechanistic structure encoded in the governing equations and the long-range temporal dependencies learned from historical data. The result is a forecasting framework that combines theoretical consistency with adaptive learning, enabling robust policy-sensitive demographic projections.

\section{Hybrid LSTM–PINN Architecture}
\label{sec 3}
Building on the policy-aware transport--reaction PDE developed in Section~\ref{sec 2}, we now introduce the hybrid LSTM--PINN architecture used to forecast India’s long-term demographic evolution. The central objective of this design is to merge the mechanistic interpretability of age-structured population dynamics with the temporal learning capabilities required to model multi-decade fertility and mortality shifts. The resulting framework enforces demographic laws through PINNs while incorporating temporal memory and policy-driven variability through LSTM networks, enabling robust, policy-sensitive forecasts over extended horizons \cite{kim2025performance, bala2024effective}.

\begin{figure}[H]
\centering
\includegraphics[width=0.85\textwidth]{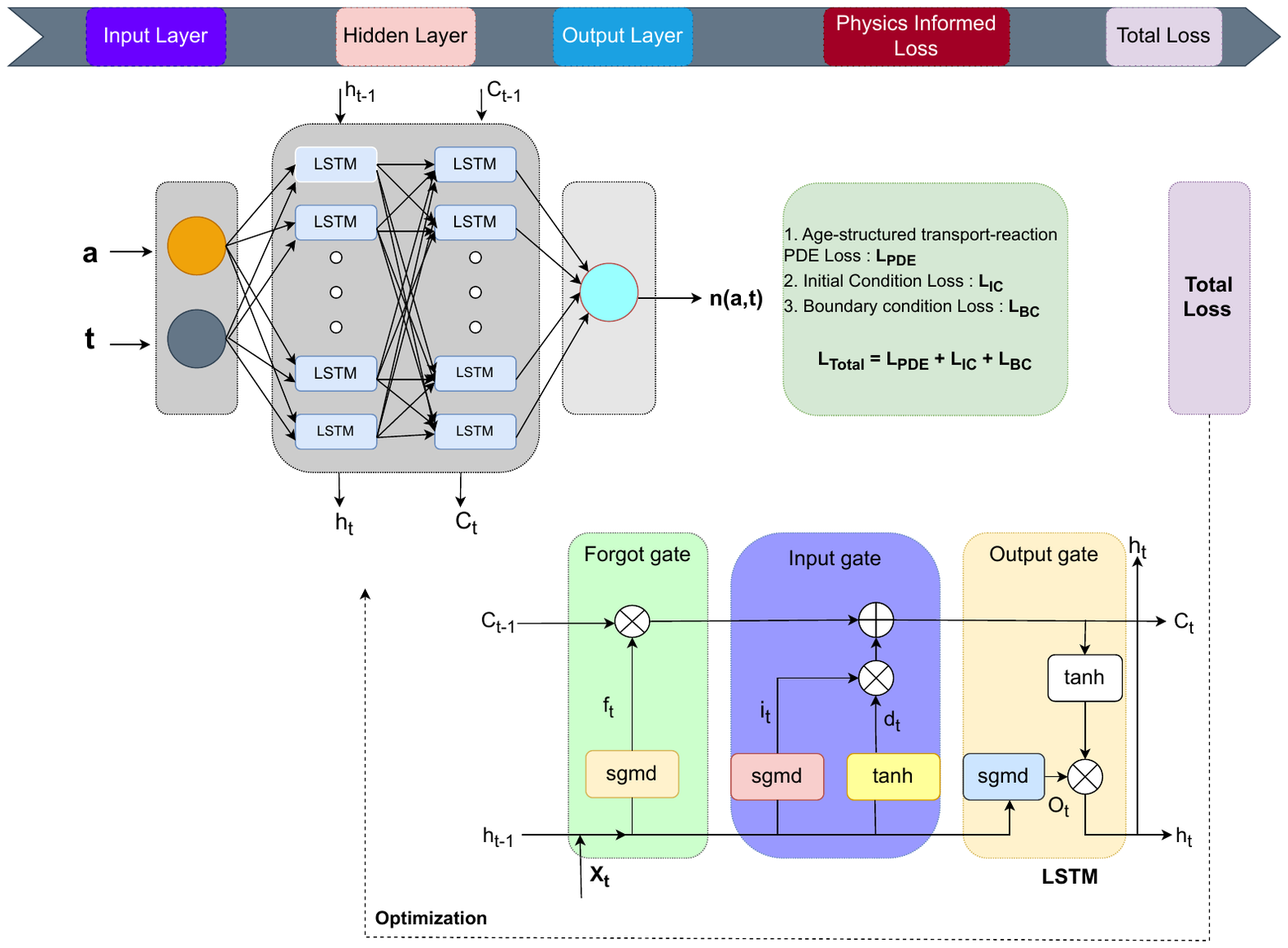}
\caption{LSTM–PINN framework integrating demographic data with PDE constraints to forecast fertility rates across age and time.}
\label{fig:pde_lstm}
\end{figure}

At the core of the architecture, a PINN approximates the age--time population density 
$n(a,t)$ by representing it through a neural surrogate $\hat{n}_\theta(a,t)$. Automatic 
differentiation is used to compute the required derivatives 
$\partial \hat{n}/\partial t$ and $\partial \hat{n}/\partial a$, allowing the network to 
evaluate the residuals of the governing PDE. Using the hybrid PDE formulation in 
Eq.~\eqref{eq:hybrid_pde}, the physics-informed residual enforced by the PINN is 
defined as:
\begin{equation}
\mathcal{R}_{\text{PDE}}(a,t) =
\frac{\partial \hat{n}}{\partial t}
+
\frac{\partial \hat{n}}{\partial a}
+
\Big[
\mu_{\text{base}}(a,t)
+
\mathcal{L}_{\text{LSTM}}^{(\mu)}
\Big]\hat{n}(a,t)
-
\Big[
B_{\text{policy}}(a,t)
+
\mathcal{L}_{\text{LSTM}}^{(B)}
\Big].
\label{eq:pinn_residual}
\end{equation}
This structure ensures that the model respects ageing dynamics, mortality processes, and 
the policy-aware birth inflow, while also incorporating LSTM-driven refinements that 
reflect long-term historical behaviour\cite{wang5101380hybrid,tao2025lstm}. The overall structure of the hybrid LSTM--PINN architecture is illustrated in 
Figure~\ref{fig:pde_lstm}

The physics constraints are reinforced by boundary and initial conditions. The initial 
condition residual enforces consistency with the observed age distribution at the starting 
year:
\begin{equation}
\mathcal{R}_{\text{IC}}(a) = 
\hat{n}(a,0) - n_{\text{data}}(a,0),
\end{equation}
while the boundary condition embeds fertility behaviour directly through the newborn 
boundary:
\begin{equation}
\mathcal{R}_{\text{BC}}(t) =
\hat{n}(0,t) -
\int_{a_{\min}}^{a_{\max}}
f(a,t)\,\hat{n}(a,t)\, da,
\end{equation}
\label{Eq. Boundary}
guaranteeing that fertility---modified by policy signals and long-term behavioural 
trends---drives the evolution of newborn cohorts in a manner consistent with demographic 
principles.

Where the PINN enforces physical consistency, the LSTM component provides the temporal 
expressiveness missing from traditional PDE-based demographic models. Demographic 
transitions in India exhibit inertia, delayed responses to socio-economic change, and 
complex generational lag effects \cite{lawal2025modeling}. To capture these, the model uses an LSTM whose input at 
each time step $t$ consists of fertility indicators, mortality summaries, policy shifts, 
and auxiliary socio-economic variables. The LSTM hidden state evolves according to:
\begin{equation}
(h_t, c_t) = \text{LSTM}(x_t, h_{t-1}, c_{t-1}),
\end{equation}
allowing the model to retain and propagate multi-decade memories.

The LSTM output is then mapped to history-aware refinements of mortality and birth 
terms:
\begin{align}
\mathcal{L}^{(\mu)}_{\text{LSTM}} &= W_\mu h_t + b_\mu, \\
\mathcal{L}^{(B)}_{\text{LSTM}} &= W_B h_t + b_B,
\end{align}
which modify $\mu(a,t)$ and $B(a,t)$ in the PDE. In this manner, mortality and fertility 
no longer evolve solely according to age and time but are dynamically shaped by temporal 
patterns learned from historical demographic trajectories. This integration is crucial for 
realistic long-term forecasting: it ensures the PDE governing population change is 
updated with memory-driven corrections rather than static or deterministic coefficients.

Training the hybrid architecture requires minimizing a combined loss function that 
balances physical consistency with data fidelity:
\begin{equation}
\mathcal{L}_{\text{total}}
=
\mathcal{L}_{\text{PDE}}
+
\lambda_{\text{IC}}\mathcal{L}_{\text{IC}}
+
\lambda_{\text{BC}}\mathcal{L}_{\text{BC}}
+
\lambda_{\text{data}}\mathcal{L}_{\text{data}},
\end{equation}
where $\mathcal{L}_{\text{PDE}}$ penalizes violations of the demographic PDE, 
$\mathcal{L}_{\text{IC}}$ and $\mathcal{L}_{\text{BC}}$ enforce initial and boundary 
conditions, and $\mathcal{L}_{\text{data}}$ encourages agreement with empirical population 
data. The hyperparameters $\lambda_{\text{IC}}, \lambda_{\text{BC}}, \lambda_{\text{data}}$ 
tune the influence of each term. All components---PINN, LSTM, and policy-aware fertility---
are trained jointly using an end-to-end differentiable process. Adam optimization is 
applied for initial convergence, followed by L-BFGS to achieve refined satisfaction of the 
PDE constraints \cite{kiyani2025optimizing}.

\begin{figure}[!ht]
    \centering
    \includegraphics[width=0.70\textwidth]{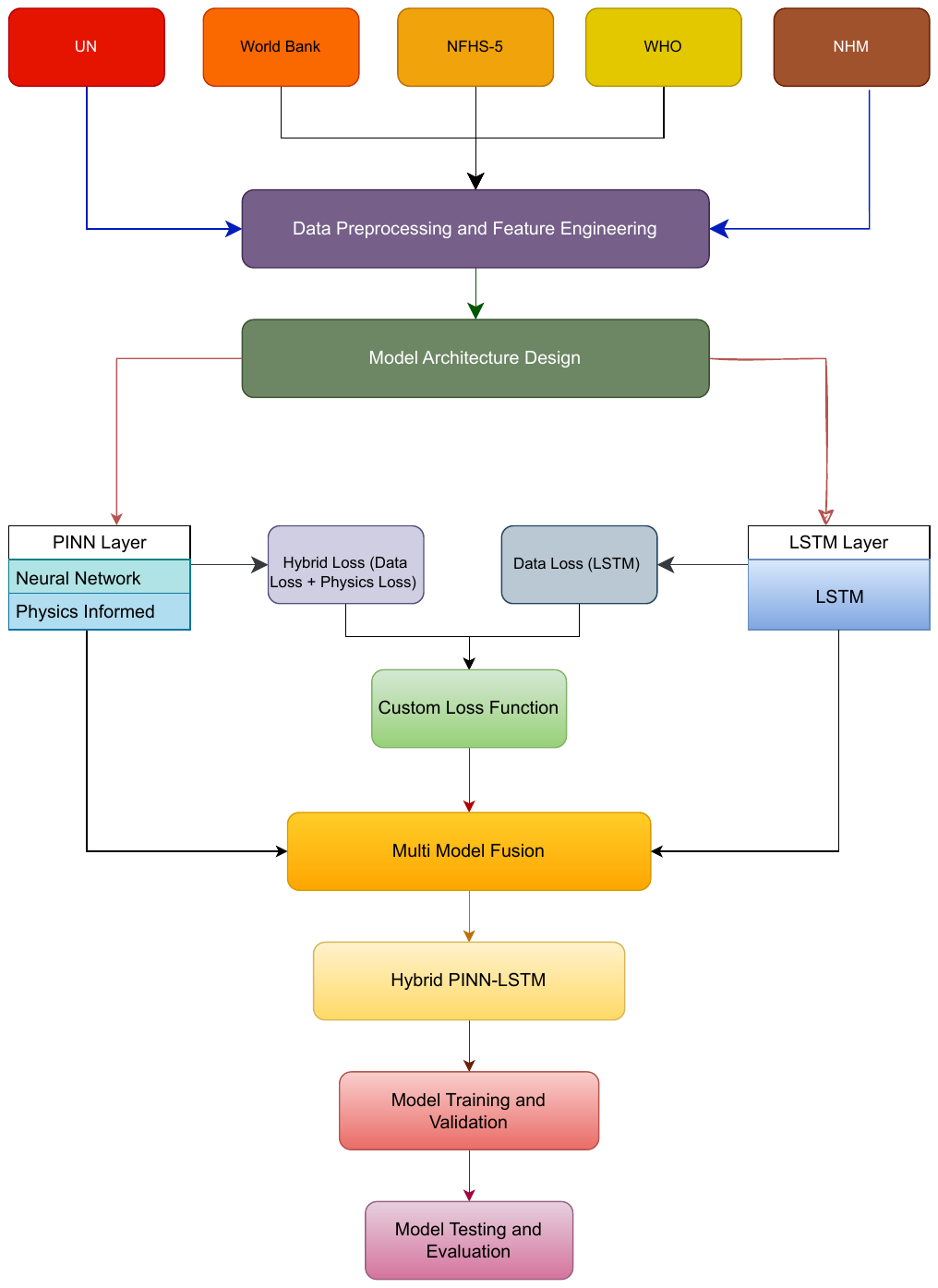}
    \caption{Workflow of the hybrid LSTM--PINN architecture. 
    The PDE model provides constraints to the PINN, whose outputs are passed through LSTM layers for long-horizon demographic forecasting.}
    \label{fig:lstm-pinn-architecture}
\end{figure}

The training workflow proceeds iteratively: collocation points in the age--time domain 
are sampled, the LSTM produces updated fertility and mortality refinements for that year, 
the PINN predicts the population field $\hat{n}(a,t)$, and physics residuals are 
evaluated. The total loss is then backpropagated through the entire architecture, updating 
both the PINN weights and the LSTM parameters simultaneously. This unified training loop 
ensures that temporal learning and physical constraints co-evolve, producing forecasts 
that are both dynamically informed and mechanistically grounded.A complete end-to-end workflow of the proposed model is shown in Figure~\ref{fig:lstm-pinn-architecture}.

Overall, the LSTM--PINN architecture offers three key advantages. First, it preserves 
\textit{physical fidelity}, ensuring that all predictions remain consistent with demographic 
conservation laws. Second, it achieves \textit{temporal sensitivity}, capturing long-term 
dependencies such as delayed policy effects and generational transitions. Third, it 
provides \textit{policy adaptivity}, enabling scenario-based simulation of India’s 
demographic future \cite{chandrasekhar2006demographic,james2011india, jain2025population} under varying fertility-policy pathways. Together, these features make 
the hybrid model especially suited for long-horizon population forecasting in complex 
socio-demographic settings like India.

By combining \textit{physical consistency}, \textit{policy awareness}, and \textit{temporal learning}, the hybrid architecture achieves both interpretability and predictive robustness, making it particularly well-suited for long-horizon demographic planning in India under multiple fertility-policy regimes \cite{chatterjee2001planning,mandelbaum1974human}.

\section{Data and Experimental Setup for India (2024–2054)}
\label{sec 4}

To ensure realistic forecasting of India's demographic evolution and labour force dynamics, the proposed hybrid LSTM-PINN model relies on official, high-quality datasets spanning multiple authoritative sources. This section describes the data collection, preprocessing, and experimental configuration used to calibrate and validate the model.

\subsection{Data Sources and Collection}

Data were collected primarily from five authoritative sources that provide comprehensive coverage of India's demographic indicators. The \textit{United Nations World Population Prospects 2022 Revision} (United Nations, 2022) supplies baseline and projected population estimates by age and sex \cite{brakman2025demography}. The \textit{World Bank World Development Indicators} (World Bank, 2024) provides macroeconomic and demographic time series \cite{world1978world}. The \textit{UNICEF Data Warehouse} (UNICEF, 2024) contributes child and maternal health indicators relevant to fertility and mortality patterns \cite{wang2021health}. The \textit{Government of India's Sample Registration System Statistical Report} (Office of the Registrar General, 2023) offers detailed vital statistics on births and deaths at state and national levels \cite{rao2021premature}. Finally, the \textit{National Family Health Survey (NFHS-5)} (Ministry of Health and Family Welfare, 2022) provides granular data on fertility preferences, contraceptive use, and reproductive health across socioeconomic strata \cite{maiti2023socioeconomic}.

Together, these sources provide the baseline and projected indicators necessary to parameterize the fertility \cite{gandotra1998fertility}, mortality \cite{suriyakala2016factors}, and migration \cite{mahapatra2020migration} components of the age-structured population dynamics equation. The data capture both historical trends and future projections aligned with India's demographic \cite{jayaraman2013demographic,chakravorty2021family} and labour force characteristics \cite{bhalla2011labour,kalyani2015unorganised}.

The following variables were extracted and harmonized for the analysis period 2024 to 2054:

\begin{itemize}
    \item \textbf{Population by Age Groups (2024 baseline):} India’s population distribution was used as an initial condition, with approximately 24.3\% in the age group 0 to 14, 68\% in the working age group 15 to 64, and 7.2\% in the elderly group 65+. The median age in 2024 is about 28.4 years (United Nations, 2022).
    
    \item \textbf{Age-Specific Fertility Rates (ASFR, ages 15 to 49):} Year-by-year fertility schedules were obtained from the UN, capturing the ongoing decline in fertility across all reproductive age groups.
    
    \item \textbf{Total Fertility Rate (TFR) Trends:} India’s TFR is projected to fall from 2.0--2.1 in 2024 to approximately 1.29 by 2050, well below replacement level, due to education, urbanization, and family planning access (NFHS-5, UN WPP).
    
    \item \textbf{Age-Specific Mortality Rates (ASMR, ages 0--100):} Mortality schedules include infant mortality, under-5 mortality, and elderly mortality, reflecting continued improvements in life expectancy. Data from UN WPP and SRS projections were interpolated for continuity.
    
    \item \textbf{Policy and Intervention Timeline:} Major population policy milestones were incorporated, such as India’s National Population Policy (2000) and Mission Parivar Vikas (2017). Although no universal two-child policy exists nationally, state-level restrictions were coded as policy-shock variables.
    
    \item \textbf{Boundary Condition Data (Births/Newborns):} In 2024, India recorded approximately 23.2 million live births, corresponding to a Crude Birth Rate (CBR) of 16.75 per 1000 population (UNICEF, World Bank).
    
    \item \textbf{Migration Data:} India’s net migration rate in 2024 is estimated at --0.456 per 1000 population, equivalent to a total net migration of about --630,830 persons (UN DESA, World Bank).
\end{itemize}

\subsection*{4.2 Experimental Setup}
The experimental design couples the policy-aware demographic PDE framework with the proposed hybrid LSTM--PINN architecture to generate long-horizon population forecasts for India. In this configuration, the transport--reaction PDE governs the core demographic processes---fertility, mortality, ageing, and migration---while the PINN component ensures adherence to the PDE structure and boundary conditions. The LSTM module complements this by learning multi-decade temporal dependencies and behavioural trends from historical demographic patterns, enabling the model to adjust fertility and mortality trajectories in a history-aware manner \cite{karniadakis2021physics, tao2025lstm}. This integrated setup ensures that the resulting forecasts reflect both mechanistic demographic constraints and realistic temporal evolution.

The baseline demographic indicators used for calibration are summarized in Table~\ref{tab:baseline}. These indicators provide the empirical foundation for setting the initial age-distribution, fertility levels, and age-specific mortality schedules for 2024, in accordance with data from UN WPP, the World Bank, and NFHS-5. The subsequent experimental configuration is structured as follows:

\begin{itemize}
    \item \textbf{Time Horizon:}  
    Forecasts are generated for the period 2024--2054 with annual resolution. This 30-year horizon captures India's ongoing demographic transition, including declining fertility, rising life expectancy, and shifts in the working-age population \cite{james2011india, subaiya2011demographics}.

    \item \textbf{Initial Conditions:}  
    The 2024 age-structured population vector $n(a,0)$ serves as the initial condition of the PDE model. This distribution is directly taken from the UN World Population Prospects (2022 Revision), ensuring consistency with internationally reported estimates \cite{desa2022world}.

    \item \textbf{Boundary Conditions:}  
    Birth inflows are computed using the product of age-specific fertility rates (ASFR) and the female population of reproductive age, following Eq. \ref{Eq. Boundary}. Mortality outflows are captured through age-specific mortality rates (ASMR), derived from UN WPP and SRS Statistical Reports. These boundary conditions define the fertility and mortality dynamics that constrain the PDE solution \cite{keyfitz1997mckendrick, mico2006age}.

    \item \textbf{Policy Variables:}  
    Policy interventions are incorporated as categorical or continuous covariates within the LSTM--PINN framework. These variables encode the timing and strength of population policies (e.g., Mission Parivar Vikas, voluntary two-child norms, state-level restrictions) and influence the fertility perturbation term $\mathcal{P}(t)$ as formulated in Section~2.1. Their inclusion enables scenario-based fertility adjustments aligned with documented policy trajectories \cite{rouyer1987political, aspalter2002population, diamond2010population}.

    \item \textbf{Training and Validation:}  
    Historical demographic data from 1990--2023 are used for pre-training and calibration of the LSTM and PINN components. This period captures key transitions in India’s fertility decline, improvements in mortality, and socio-economic changes observed across NFHS surveys and UN projections. Forecasts generated for 2024 onward are validated against UN medium-variant projections and historical consistency criteria to ensure plausibility and structural fidelity \cite{bernardi2025heterogeneously}.
\end{itemize}

This configuration enables the hybrid LSTM--PINN model to leverage robust empirical foundations while maintaining strict adherence to demographic theory, ultimately producing stable and interpretable long-term forecasts for India's population under varying policy scenarios.

\begin{table}[!ht]
\centering
\caption{Baseline demographic indicators for India (2024). Sources: UN WPP, World Bank, NFHS-5.}
\label{tab:baseline}
\begin{tabular}{l c}
\hline
Median Age & 28.4 years \\
Population (0--14) & 24.3\% \\
Population (15--64) & 68.0\% \\
Population (65+) & 7.2\% \\
Total Fertility Rate (TFR) & $\approx$ 2.0 \\
Live Births & 23.2 million \\
Crude Birth Rate (CBR) & 16.75 per 1000 \\
Net Migration (2024) & --630,830 \\
\hline
\end{tabular}
\end{table}

\section{Results and Policy Implications}
\label{sec 5}
This section presents the forecasting outcomes generated by the proposed LSTM--PINN framework under multiple fertility-policy scenarios for India. The results illustrate how the model reproduces age-structured demographic dynamics, captures long-range temporal dependencies, and responds to alternative policy conditions embedded in the fertility and mortality functions. We report population density trajectories, loss convergence behaviour, and comparative scenario outcomes, followed by an examination of their broader implications for India’s demographic transition between 2024 and 2054.
\subsection{Baseline PINN Framework with Deterministic Birth Integration}

The first set of experiments employed the refined PINN model, incorporating deterministic quadrature for birth boundary conditions and training on GPU \cite{escapil2023h,sharma2022accelerated} with a learning-rate scheduler. The model demonstrates stable convergence across all three fertility scenarios. 

\begin{itemize}
    \item \textbf{Baseline fertility (TFR $\approx 2.0$):} Training loss declined smoothly from an initial value of $\sim 8.7 \times 10^{-1}$ to $\sim 4.7 \times 10^{-2}$ by epoch 9000.
    \item \textbf{Declining fertility (TFR $\to 1.6$):} The total loss stabilized near $\sim 5.4 \times 10^{-2}$, reflecting the sensitivity of long-term population size to marginal fertility reduction. 
    \item \textbf{Policy boost (TFR $\to 2.2$):} The model stabilized at $\sim 2.8 \times 10^{-2}$, highlighting the gains in younger cohorts under fertility incentives. 
\end{itemize}

\begin{figure}[H]
    \centering
    \includegraphics[width=0.8\textwidth]{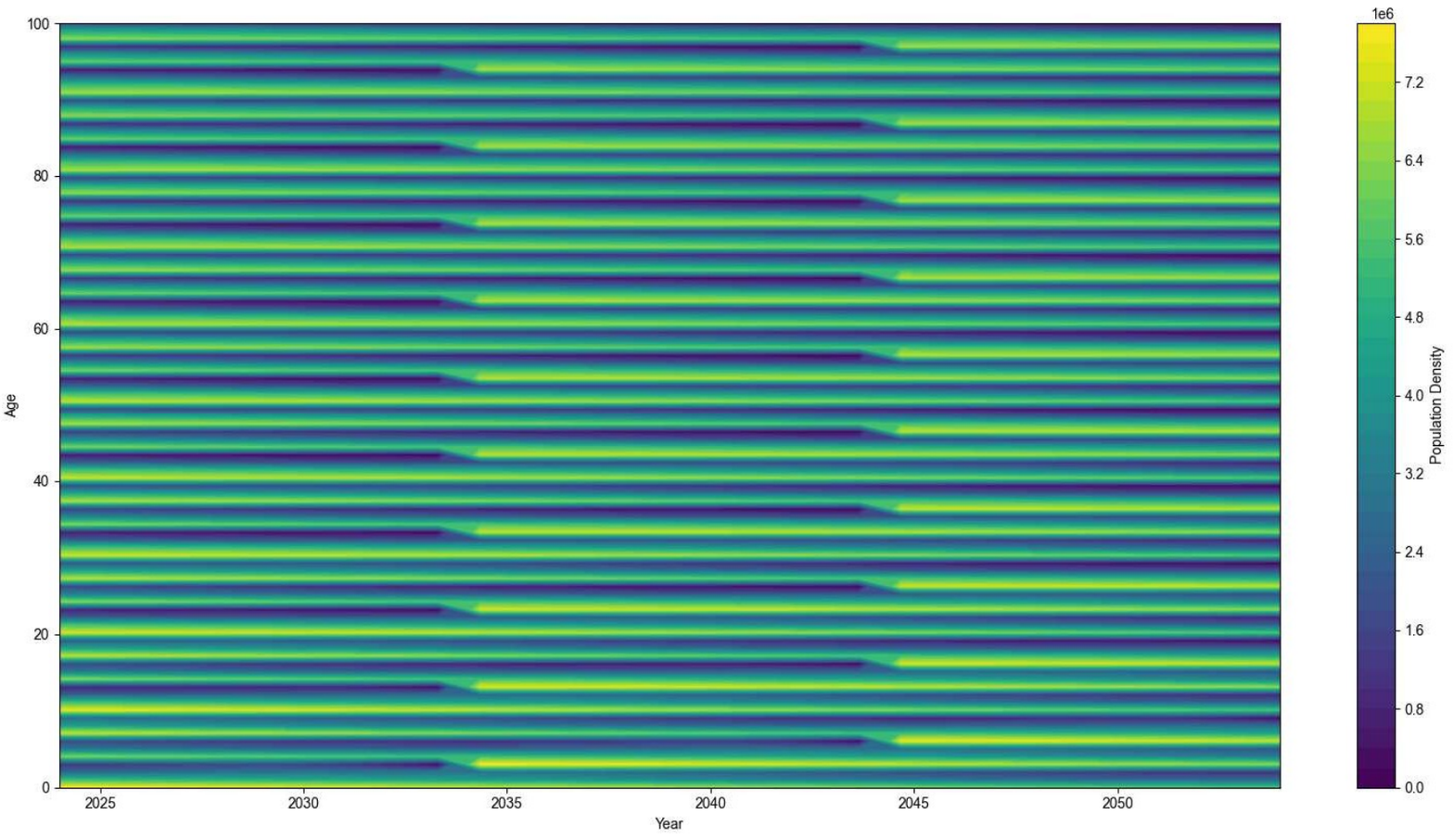}
    \vspace{-2em}
    \caption{India population projections (2024--2054) under Baseline fertility (TFR $\approx 2.0$) using the PINN model.}
    \label{Baselin}
\end{figure}

\begin{figure}[H]
    \centering
    \includegraphics[width=0.8\textwidth]{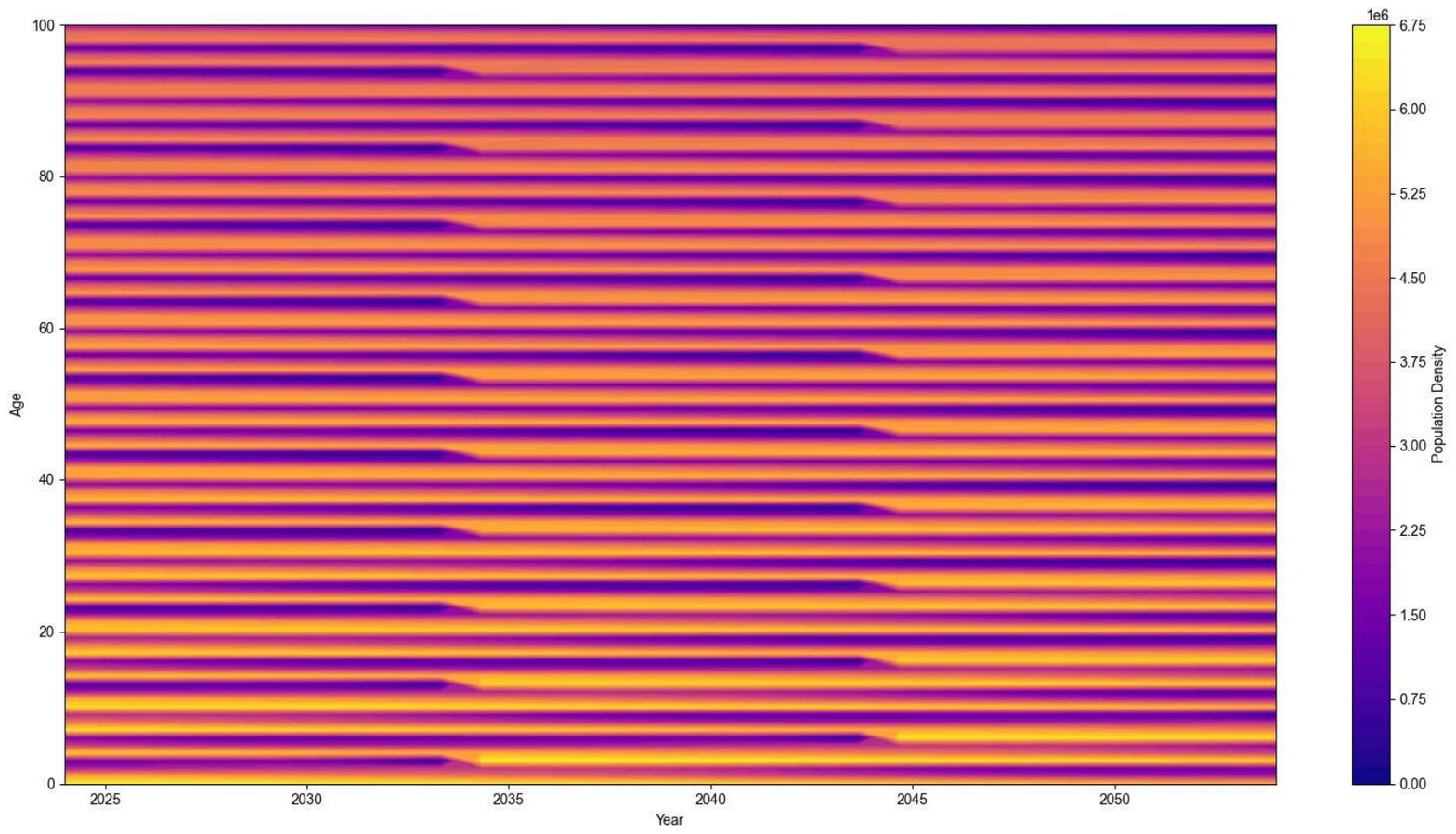}
    \vspace{-2em}
    \caption{India population projections (2024--2054) under declining fertility (TFR $\to 1.6$) using the PINN model.}
    \label{Declinin}
\end{figure}

\begin{figure}[H]
    \centering
    \includegraphics[width=0.8\textwidth]{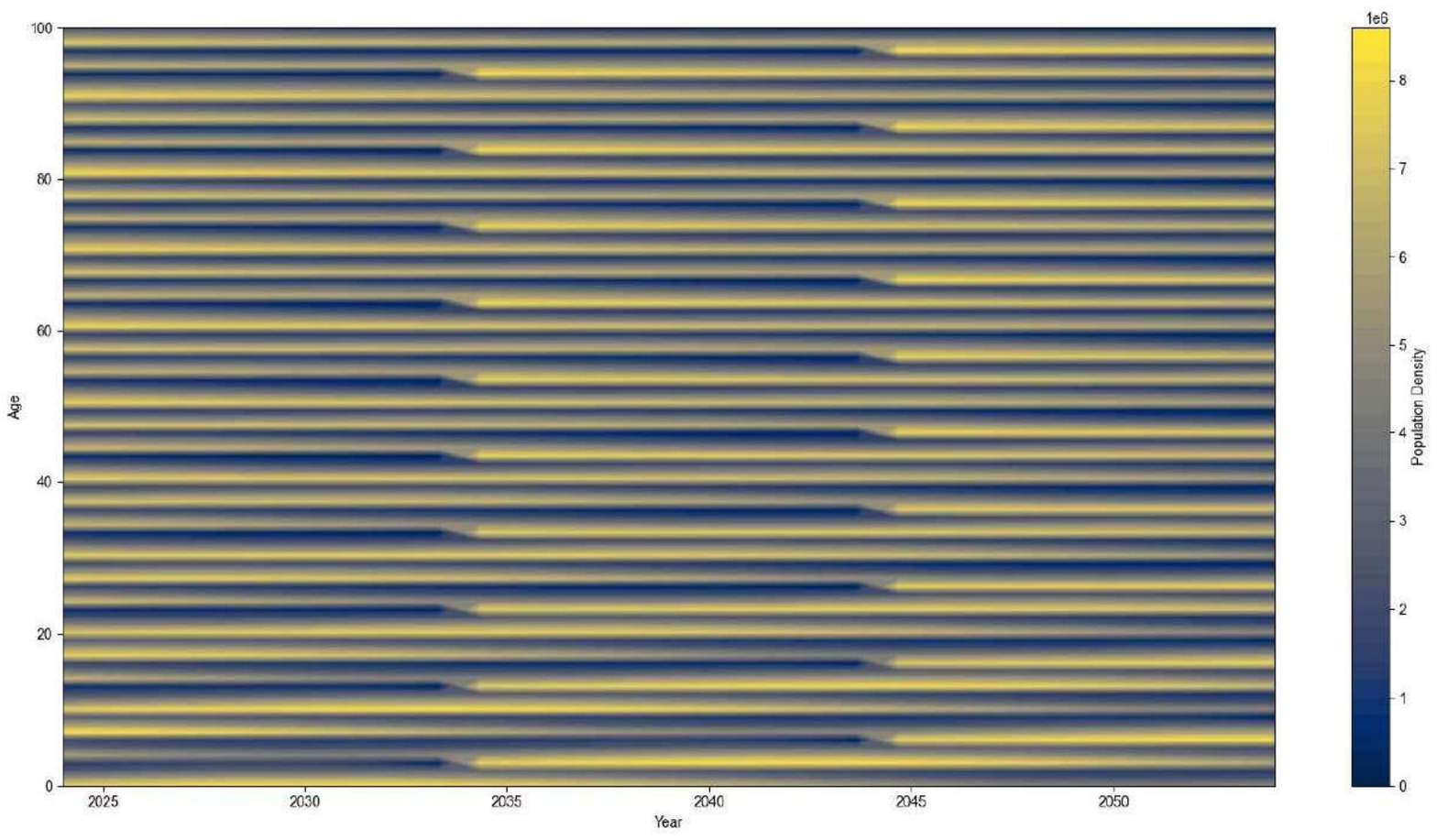}
    \vspace{-2em}
    \caption{India population projections (2024--2054) under Policy boost (TFR $\to 2.2$) using the PINN model.}
    \label{Policy}
\end{figure}

\begin{figure}[H]
    \centering
    \includegraphics[width=0.8\textwidth]{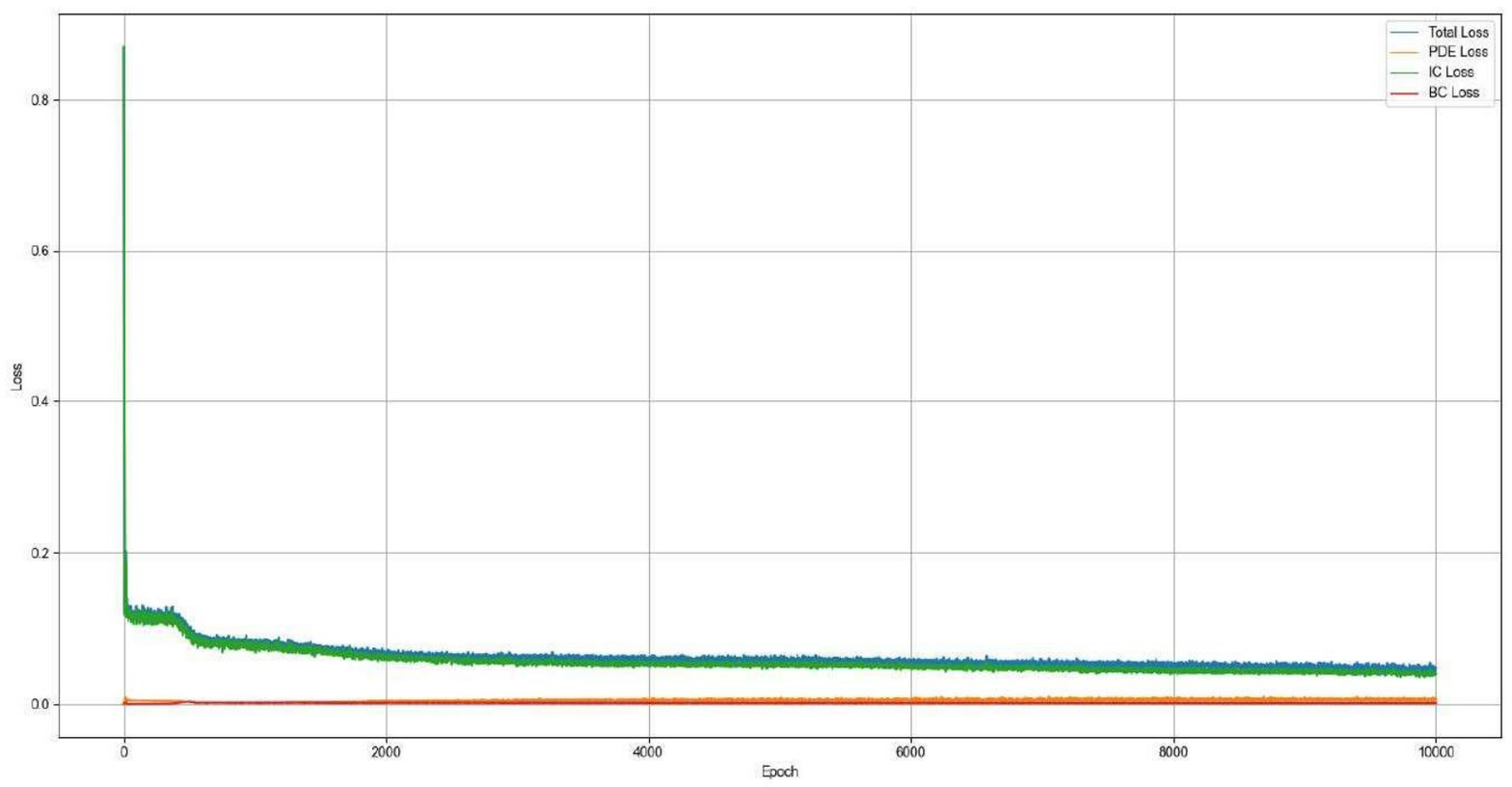}
    \vspace{-2em}
    \caption{Training loss curve for the Baseline fertility scenario (PINN).}
    \label{loss Baseline}
\end{figure}

\begin{figure}[H]
    \centering
    \includegraphics[width=0.8\textwidth]{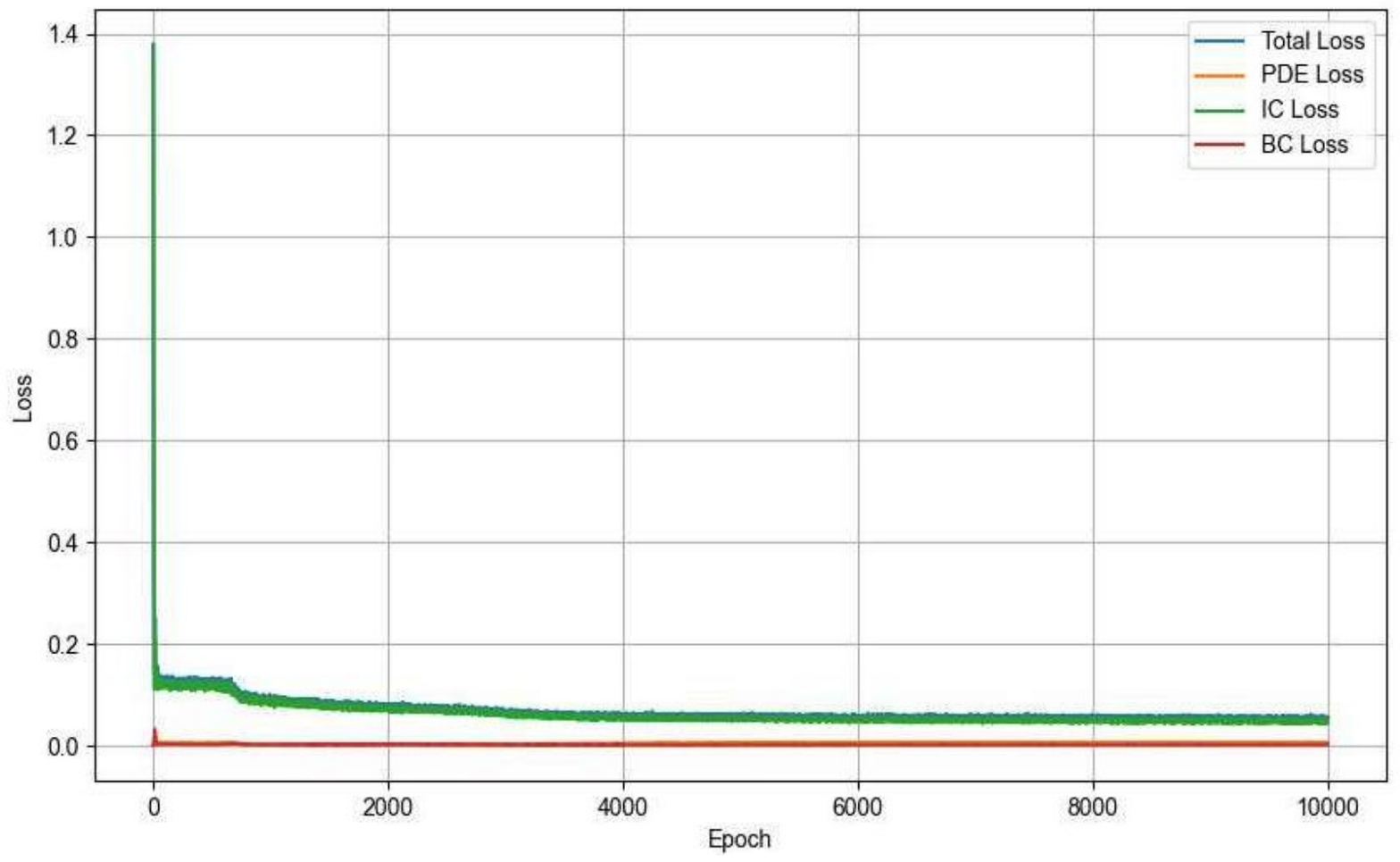}
    \vspace{-2em}
    \caption{Training loss curve for the Declining fertility scenario (PINN).}
    \label{loss Declining}
\end{figure}

\begin{figure}[H]
    \centering
    \includegraphics[width=0.8\textwidth]{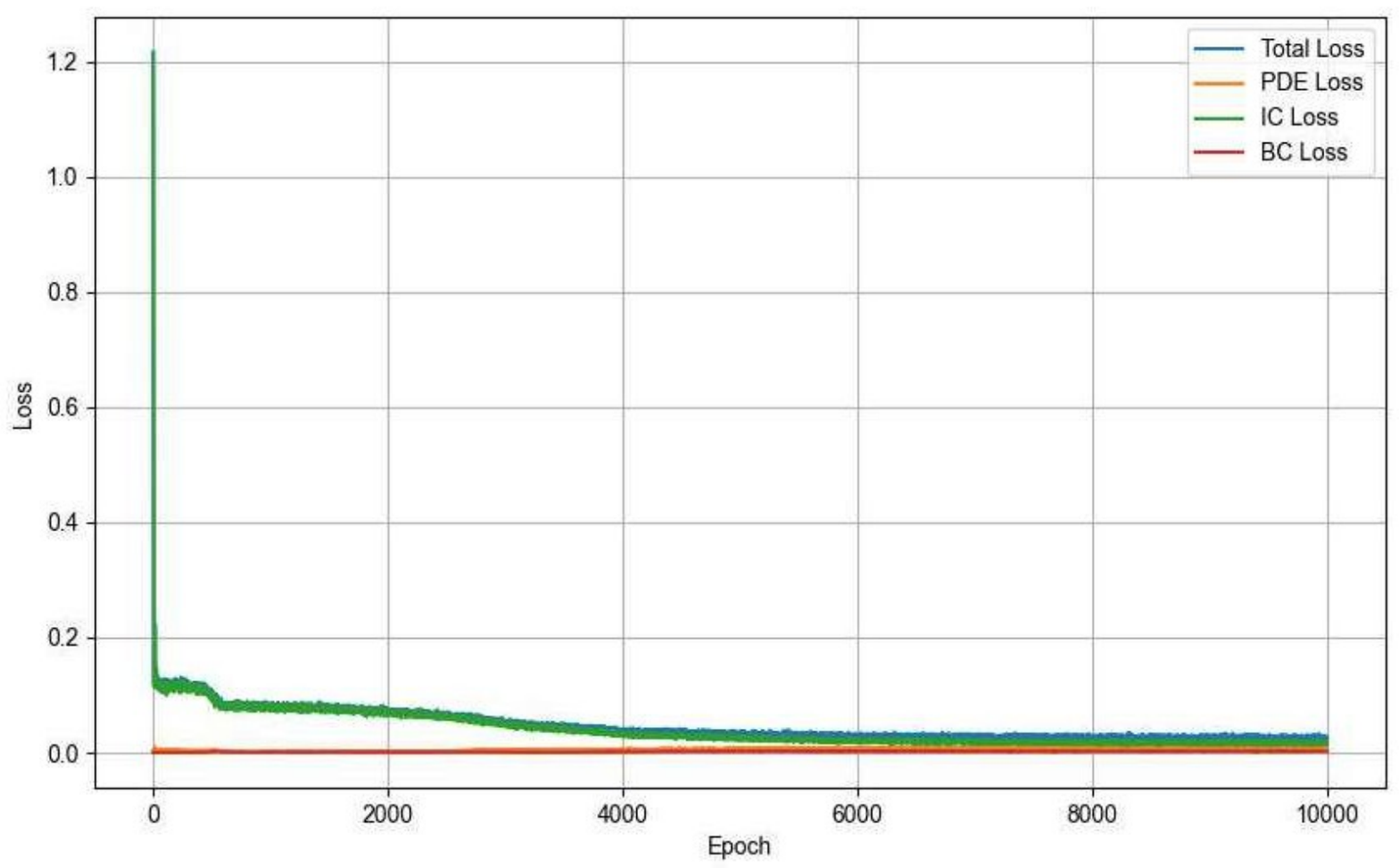}
    \vspace{-2em}
    \caption{Training loss curve for the Policy boost scenario (PINN).}
    \label{loss Policy}
\end{figure}

Figure \ref{Baselin} illustrates the age–time population density under the baseline (TFR $\approx 2.0$), showing stable but gradually narrowing cohorts.  The model captures India’s current demographic trajectory, with gradual narrowing of younger cohorts and stable movement of age bands over time. This serves as an essential benchmark for evaluating alternative fertility-path assumptions.
Figure  \ref{Declinin} shows the scenario with declining fertility (TFR $\to 1.6$), where the base of the age structure contracts significantly over time.  
This highlights the risk of shrinking workforce and accelerated population ageing, directly aligning with the study’s policy relevance. Figure \ref{Policy} presents the policy boost case (TFR $\to 2.2$), sustaining broader bases and larger young cohorts.  
This demonstrates how pronatalist policies can delay ageing and maintain workforce size, fulfilling the objective of policy-aware projections. Figure \ref{loss Baseline} plots the training loss for the baseline scenario, where total, PDE, IC, and BC losses all converge steadily.  
This validates that the PINN model respects the governing PDE and boundary conditions, ensuring reliable projections. Figure \ref{loss Declining} shows the loss components under the declining fertility scenario, with consistent convergence across PDE, IC, and BC terms.  
This confirms that the model robustly learns altered fertility dynamics while maintaining mechanistic consistency. Figure \ref{loss Policy} reports the training loss for the policy boost case, converging to low residuals despite sharper boundary forcing.  
This indicates the model’s robustness in integrating policy shocks into the demographic PDE, satisfying the objective of scenario testing.
Overall, the baseline PINN results confirm that the model accurately reproduces age-structured population dynamics and is capable of representing the demographic consequences of fertility shifts under stable mechanistic constraints. 

\begin{figure}[htbp]
    \centering
    \begin{subfigure}[b]{0.48\textwidth}
        \centering
        \includegraphics[width=\textwidth]{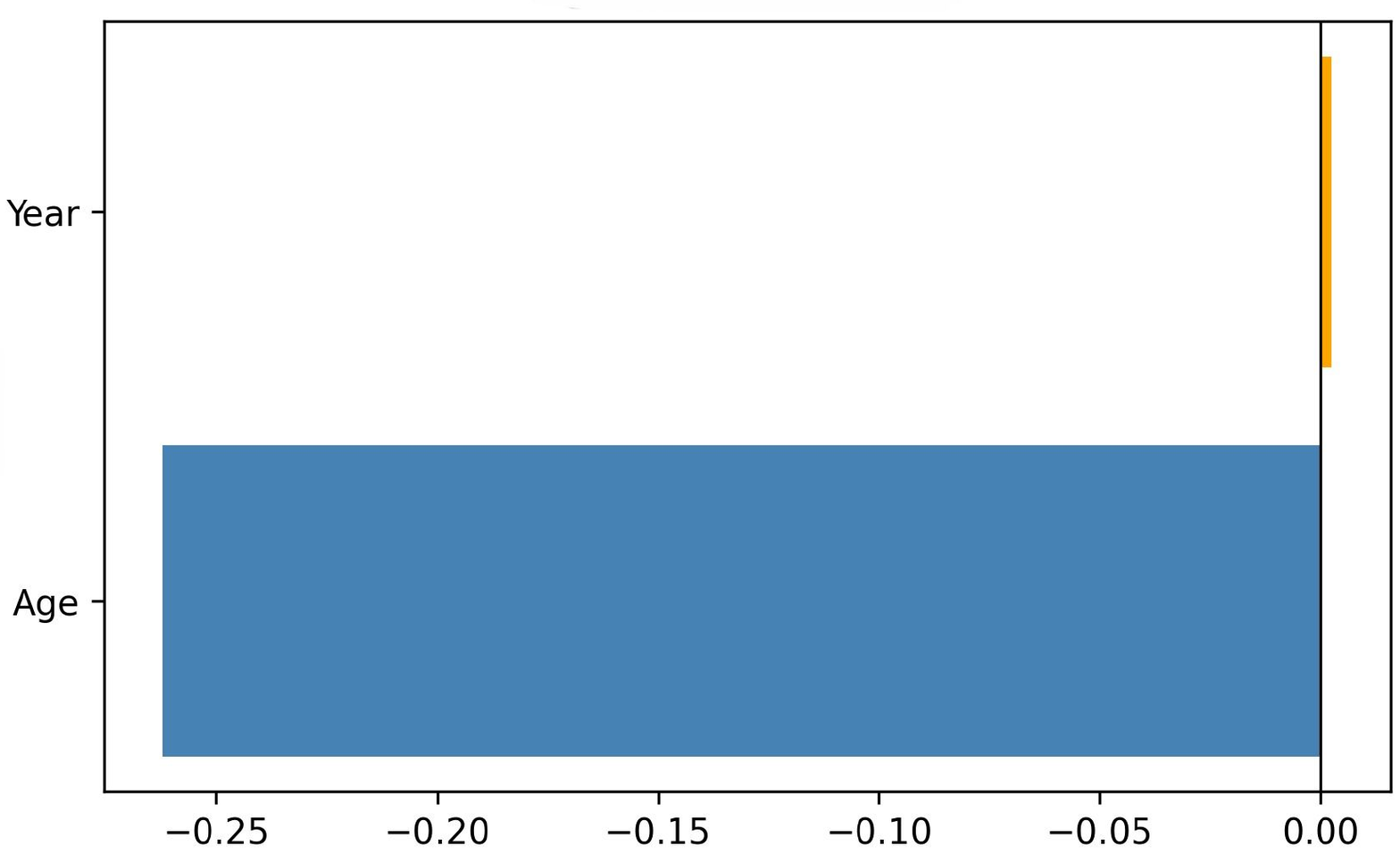}
        \caption{Age = 30, Year = 2039}
    \end{subfigure}
    \hfill
    \begin{subfigure}[b]{0.48\textwidth}
        \centering
        \includegraphics[width=\textwidth]{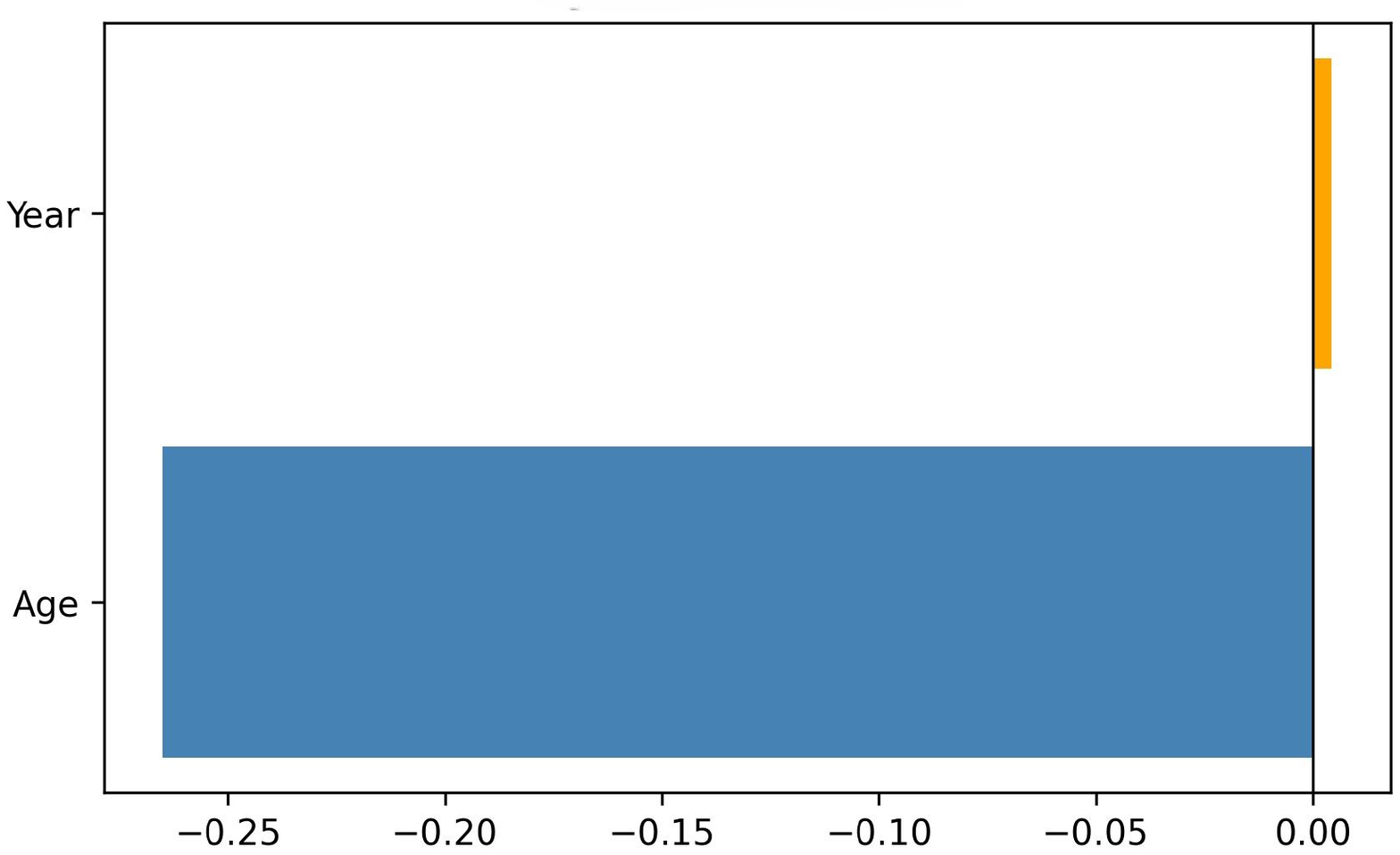}
        \caption{Age = 50, Year = 2045}
    \end{subfigure}

   \caption{LIME explanations for PINN demographic forecast}
    \label{fig:lime_pinn}
\end{figure}

To assess the interpretability of the PINN predictions, a LIME explanation \cite{ribeiro2016should} is provided in Figure~\ref{fig:lime_pinn}. The plot highlights the relative influence of key input features on the model’s local forecast. Age (sky-blue bar) exhibits a strong negative contribution, indicating that higher ages reduce the predicted population density, consistent with natural mortality progression. In contrast, the Year variable (orange bar) contributes only marginally and positively, reflecting modest temporal adjustments across adjacent forecast years. This pattern aligns well with demographic theory, where age is the dominant determinant of survival dynamics, while short-term temporal changes exert minimal influence.\\
The LIME output therefore reinforces that the PINN behaves in an interpretable and demographically coherent manner, demonstrating that the model’s decisions are driven by meaningful population factors rather than opaque black-box effects.


\vspace{7 cm}

\subsection{Hybrid LSTM–PINN with Policy-Aware Fertility Embedding}

The second model integrated an LSTM-enhanced PINN architecture to better capture temporal dependencies in fertility and mortality patterns. 

\begin{itemize}
    \item \textbf{Baseline (no new policy):} Total loss decreased from $\sim 1.15 \times 10^{0}$ at initialization to $\sim 2.7 \times 10^{-2}$ by epoch 9500.
    \item \textbf{Two-child voluntary policy (2024 onward):} Training stabilized near $\sim 2.6 \times 10^{-2}$, with narrowing of the reproductive-age band. 
    \item \textbf{Enhanced family planning (Mission Parivar Vikas):} The model converged to $\sim 2.3 \times 10^{-2}$, revealing contraction at the base of the age pyramid and accelerated demographic transition. 
\end{itemize}

\begin{figure}[H]
    \centering
    \includegraphics[width=0.8\textwidth]{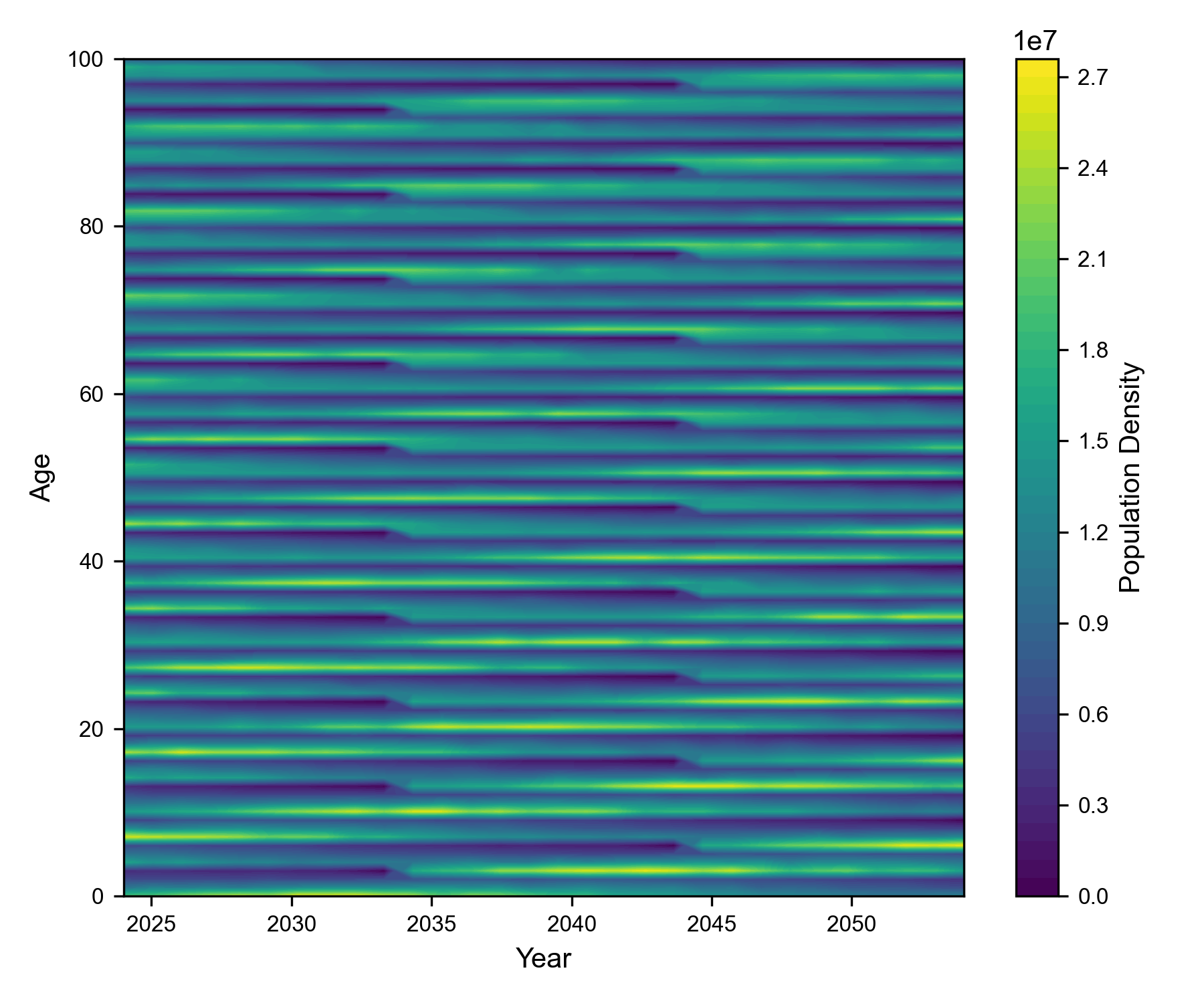}
    \vspace{-1em}
    \caption{India population projections (2024--2054) under Baseline fertility} using the hybrid LSTM–PINN model.
    \label{Baselin 2}
\end{figure}

\begin{figure}[H]
    \centering
    \includegraphics[width=0.7\textwidth]{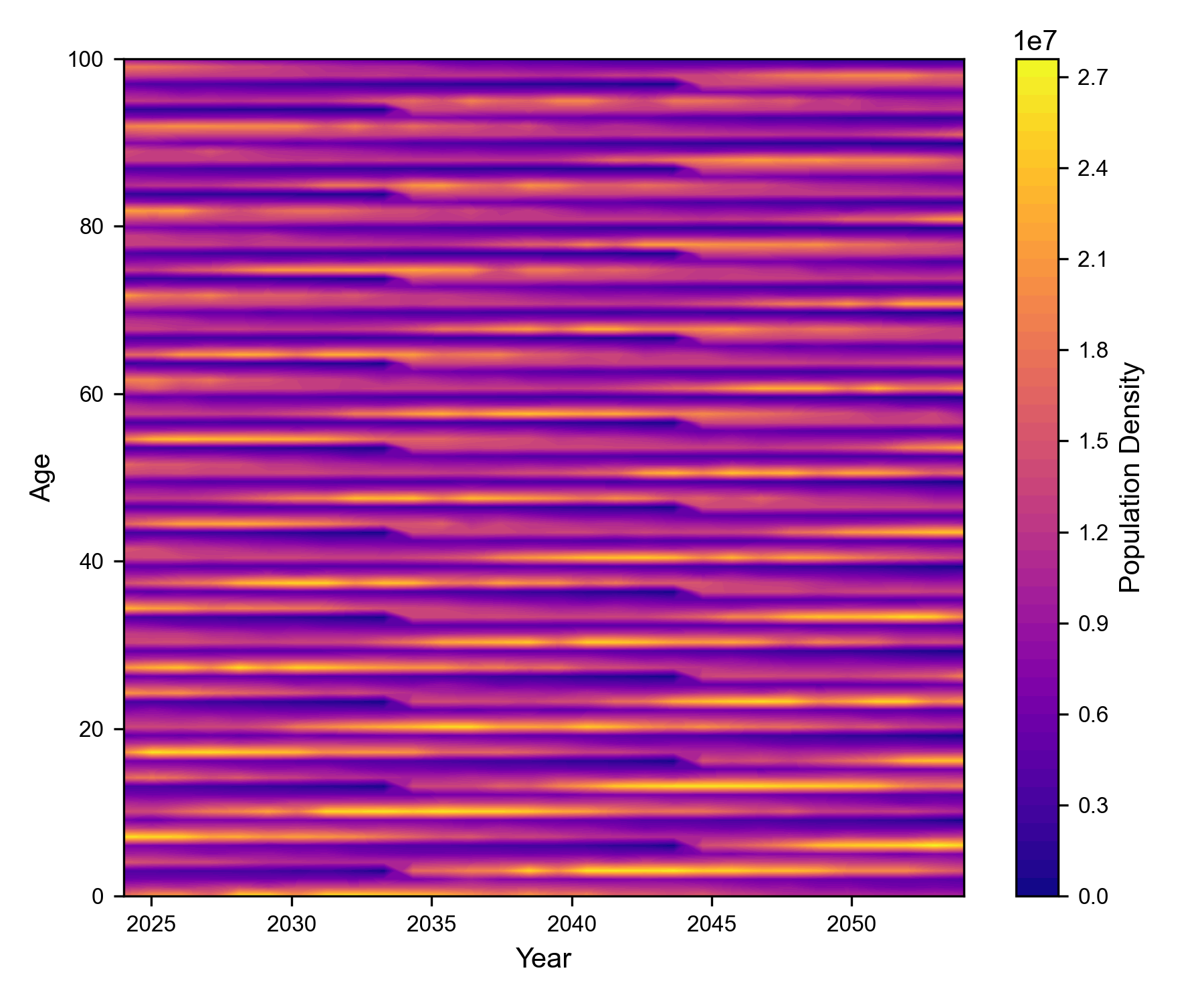}
    \vspace{-1em}
    \caption{India population projections (2024--2054) under the Two-child voluntary policy (2024)} using the hybrid LSTM–PINN model.
    \label{Two-child voluntary}
\end{figure}

\begin{figure}[H]
    \centering
    \includegraphics[width=0.7\textwidth]{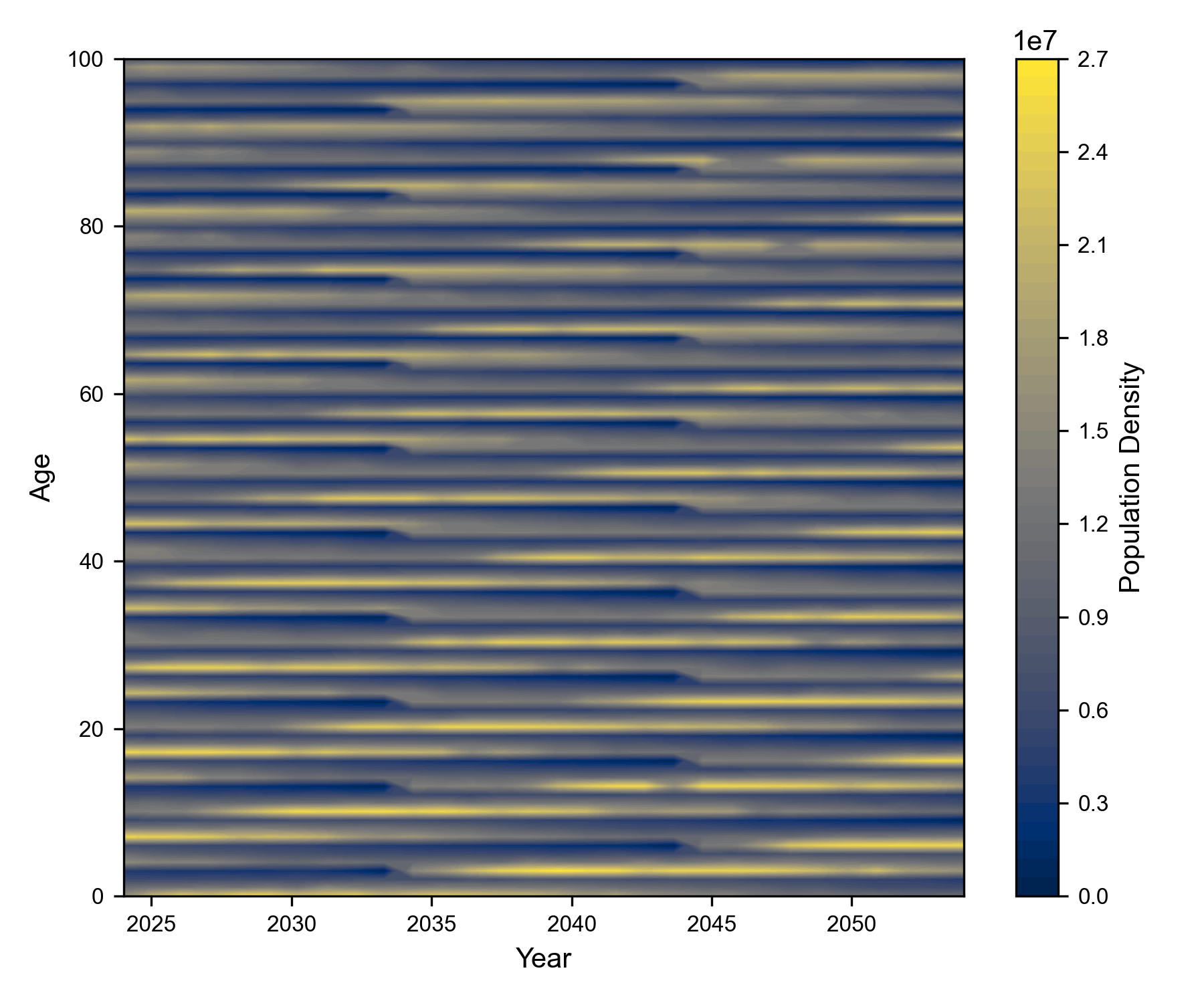}
    \caption{India population projections (2024--2054) under Enhanced family planning (Mission Parivar Vikas) using the hybrid LSTM–PINN model.}
    \label{Enhanced family}
\end{figure}

\begin{figure}[H]
    \centering
    \includegraphics[width=1.0\textwidth]{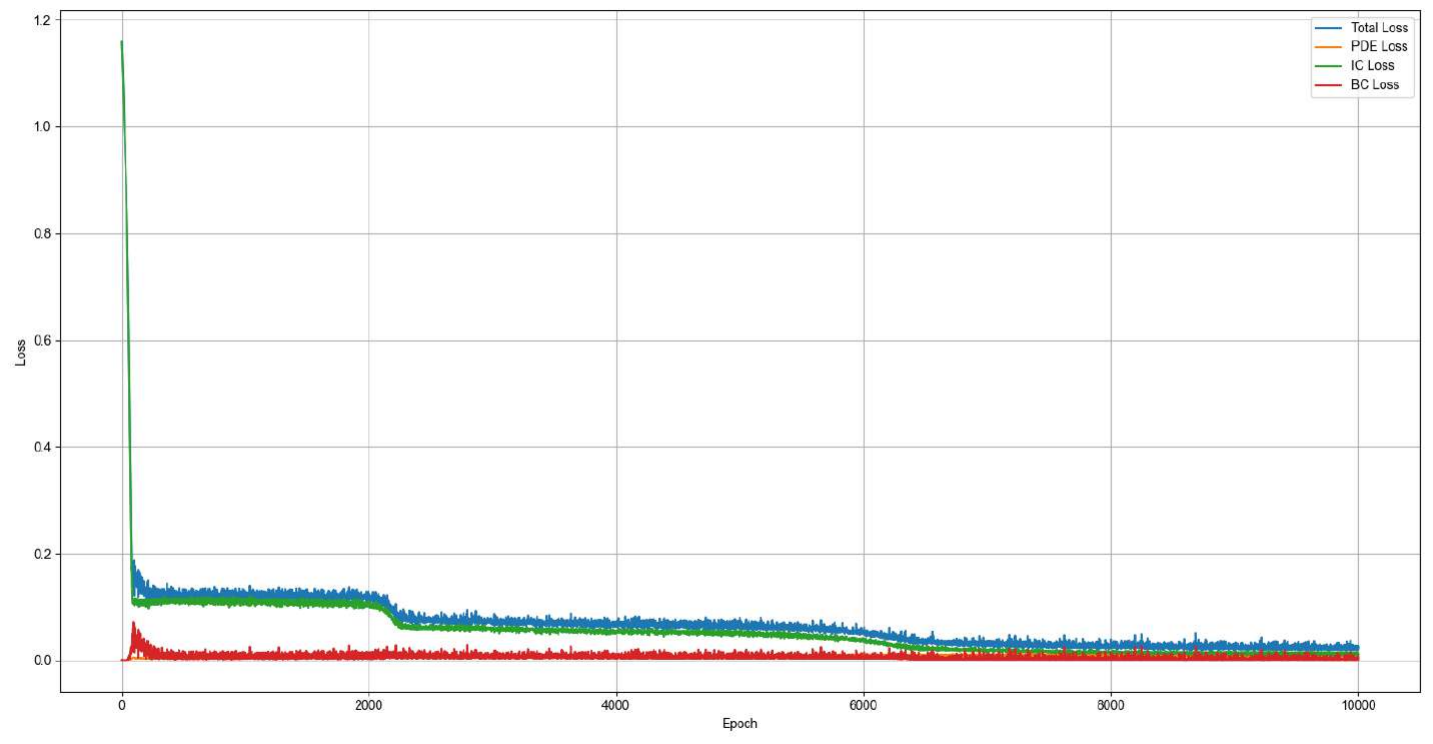}
    \vspace{-6em}
    \caption{Training loss curve for the Baseline fertility scenario (hybrid LSTM–PINN).}
    \label{Loss Baselin 2}
\end{figure}

\begin{figure}[H]
    \centering
    \includegraphics[width=1.0\textwidth]{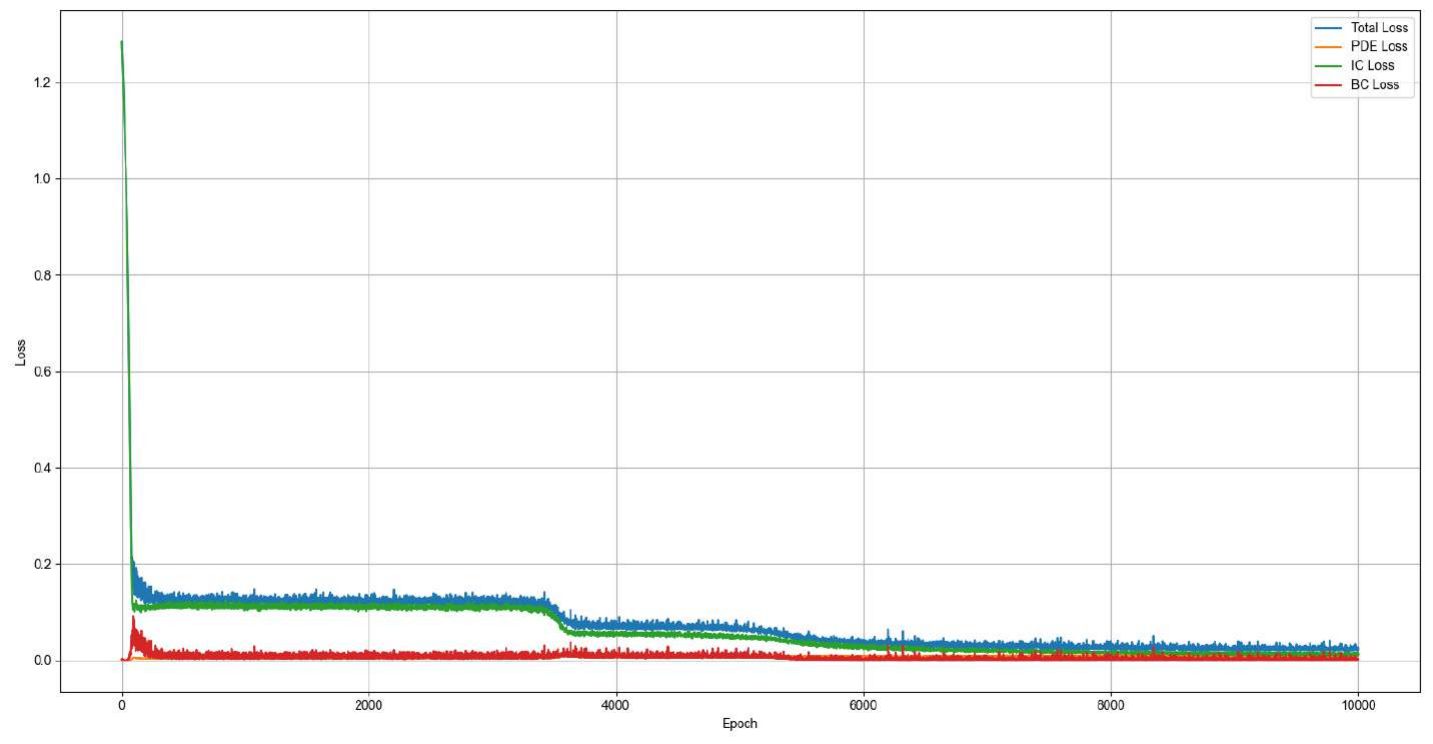}
    \vspace{-6em}
    \caption{Training loss curve for the Two-child voluntary policy scenario (hybrid LSTM–PINN)}.
    \label{Loss Two-child volunt}
\end{figure}

\begin{figure}[H]
    \centering
    \includegraphics[width=1.0\textwidth]{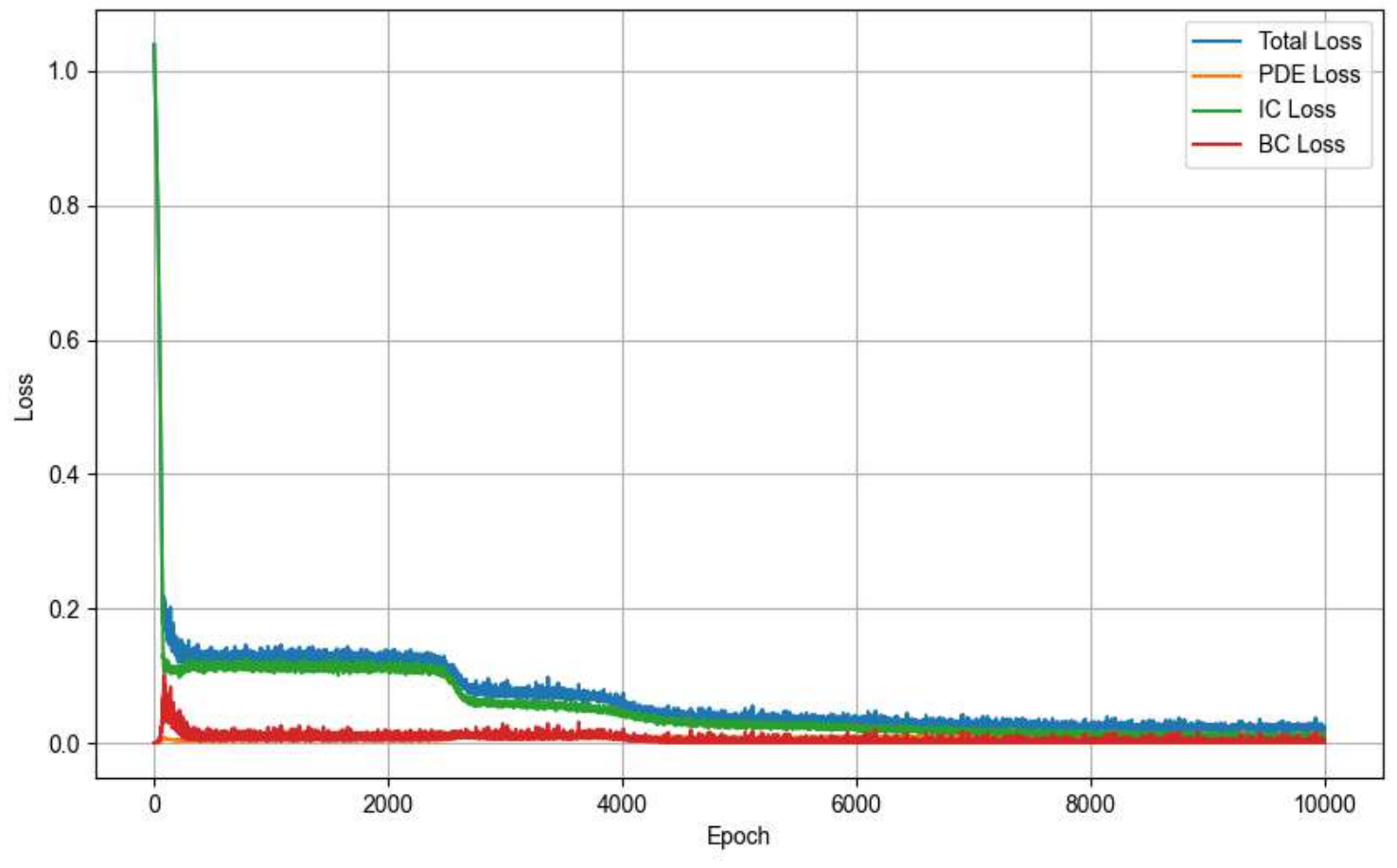}
    \vspace{-4em}
    \caption{Training loss curve for the Enhanced family planning scenario (hybrid LSTM–PINN).}
    \label{Loss Enhanced family}
\end{figure}

The hybrid model produced smoother age–time trajectories and reduced loss magnitudes compared to the baseline PINN, confirming the advantage of embedding temporal memory.
Figure~\ref{Baselin 2} shows the age–year population density for the LSTM–PINN baseline scenario (2024–2054), with diagonal cohort bands and a sustained working-age bulge.  This confirms the model’s ability to produce India-specific, mechanistically consistent baseline projections for policy comparison. Figure~\ref{Two-child voluntary} shows population density under the voluntary two-child scenario, with a narrowing base and smaller cohorts post-2024.  
This illustrates how moderate fertility policies propagate through the age structure, informing workforce and dependency planning. Figure~\ref{Enhanced family} presents the enhanced family-planning scenario, where stronger fertility reduction yields a visibly contracted base and accelerated ageing.  
This highlights the demographic trade-off of aggressive fertility policies—reduced births now, faster ageing later—central to policy decisions. Figure~\ref{Loss Baselin 2} plots training loss components for the LSTM–PINN baseline; total loss and PDE/BC residuals converge to low values.  
This validates that model forecasts are physically consistent (PDE + BC satisfied) and numerically stable.
Figure~\ref{Loss Two-child volunt} shows loss convergence under the two-child policy; losses stabilize despite boundary adjustments due to policy changes.  
This demonstrates the model’s robustness in learning policy-driven fertility functions while maintaining mechanistic constraints. Figure~\ref{Loss Enhanced family} shows training diagnostics for the enhanced-planning scenario, converging to low residuals after transient BC adjustments.  

Across all cases, the LSTM--PINN achieves lower loss magnitudes and smoother age--time trajectories than the baseline PINN, indicating the benefit of explicitly modelling temporal dependencies. The results also confirm that the hybrid model effectively integrates sharper policy shocks into the population PDE and yields reliable projections.

\begin{figure}[htbp]
    \centering
    \begin{subfigure}[b]{0.45\textwidth}
        \centering
        \includegraphics[width=\textwidth]{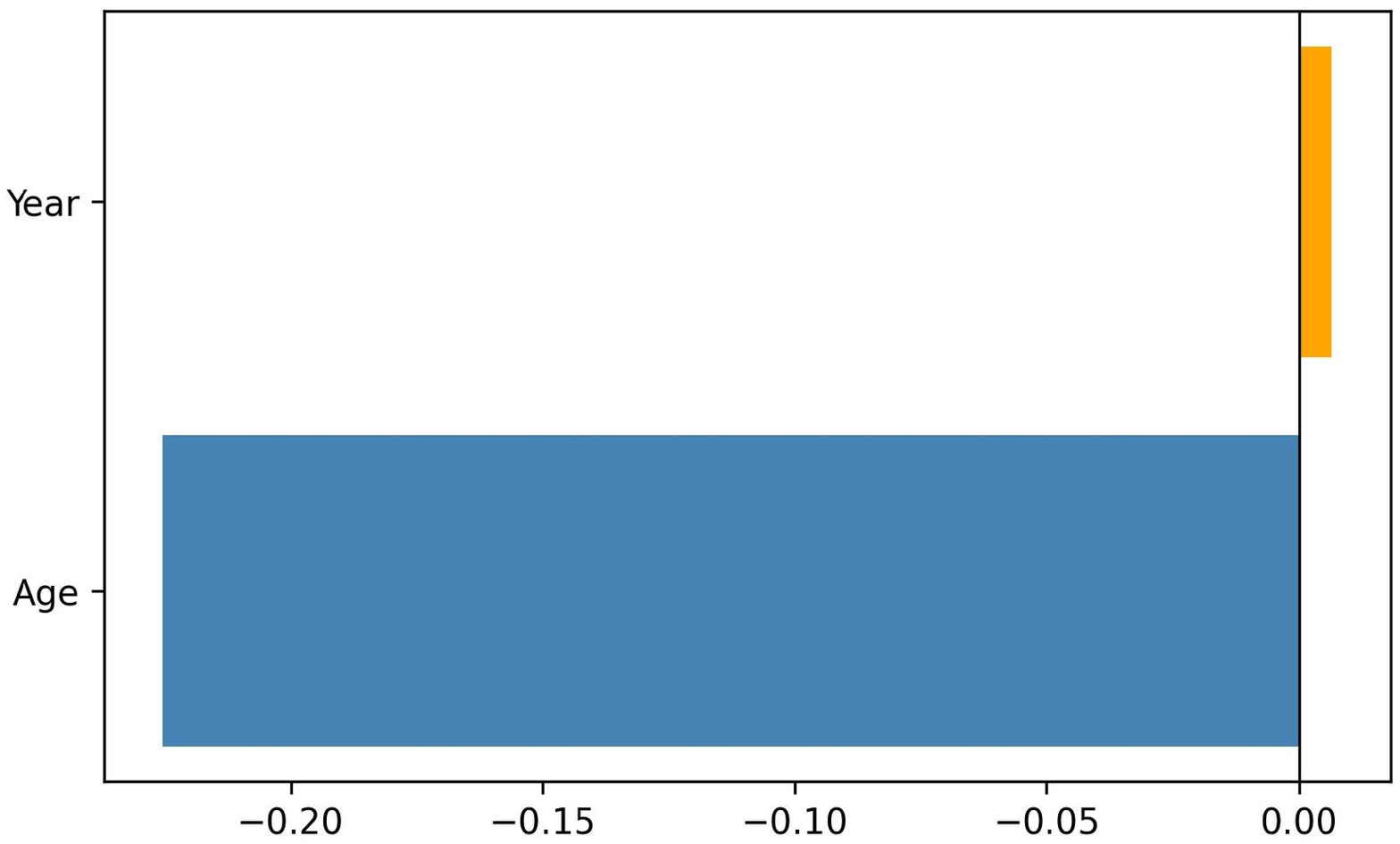}
        \caption{Age = 30, Year = 2039}
    \end{subfigure}
    \hfill
    \begin{subfigure}[b]{0.45\textwidth}
        \centering
        \includegraphics[width=\textwidth]{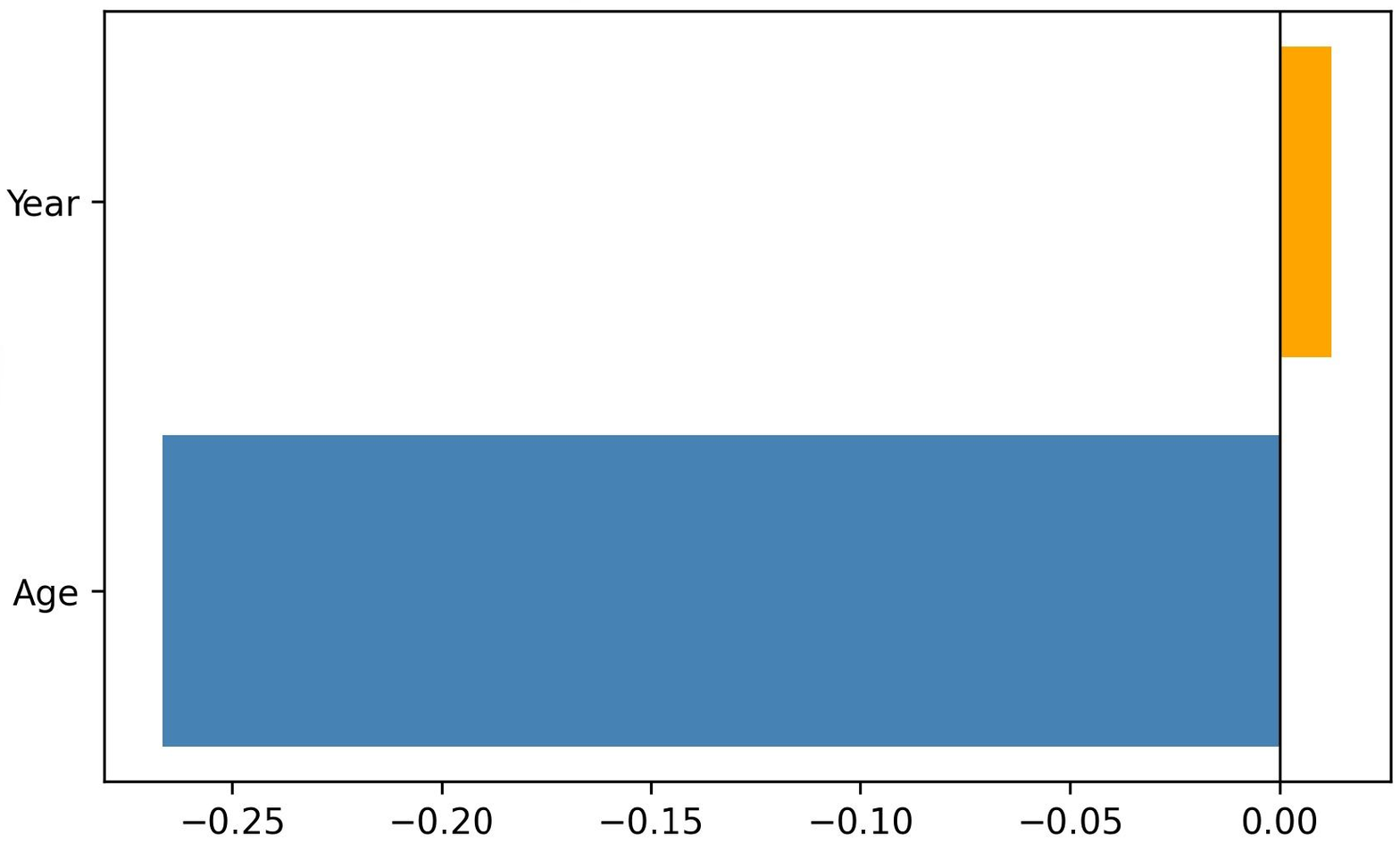}
        \caption{Age = 70, Year = 2054}
    \end{subfigure}

   \caption{LIME explanations for hybrid LSTM-PINN demographic forecast}
    \label{fig:lime_ Lstm pinn}
\end{figure}

Model interpretability for the LSTM--PINN is assessed using the LIME explanation \cite{ribeiro2016should} shown in Figure \ref{fig:lime_ Lstm pinn}. As in the baseline PINN, Age (blue bar) exhibits a strong negative contribution, indicating that higher ages reduce predicted population density, consistent with demographic survivorship patterns. The Year variable (orange bar) exerts only a minor positive influence, reflecting modest year-to-year adjustments relative to age structure. This behaviour aligns with demographic theory and confirms that the hybrid model remains interpretable: its predictions are driven by meaningful demographic factors rather than opaque temporal correlations. The LIME results therefore validate the transparency and reliability of the LSTM--PINN framework for long-term population forecasting.

\vspace{5cm}

\subsection{Policy Implications and Model Comparison}

The combined findings from the PINN and LSTM--PINN models provide several policy-relevant insights into India’s demographic trajectory over 2024--2054. Across all scenarios, the models show that fertility patterns exert the strongest long-term influence on the age structure, especially on the balance between young and working-age cohorts \cite{bloom2001economic,bloom2003demographic}. The declining-fertility scenario illustrates a rapid contraction in younger populations, signalling risks of future labour shortages, rising old-age dependency, and accelerated demographic ageing. These outcomes underline the importance of maintaining a balanced fertility level to support economic productivity and social welfare systems.

Moderate interventions, such as voluntary two-child norms, yield noticeable but manageable reductions in young cohorts. The projections indicate that such policies would gradually reshape the age distribution without causing abrupt demographic shocks. In contrast, more aggressive fertility-reduction initiatives—such as enhanced family planning—produce a sharper contraction at the base of the age pyramid. While these policies may achieve short-term population stabilization goals, the long-term consequence is a substantially older population, potentially intensifying fiscal pressure on healthcare, pensions, and social support systems.

The policy-boost (higher fertility) scenario demonstrates the opposite dynamic: broader base cohorts help sustain the working-age population over the forecast horizon, delaying ageing and supporting workforce continuity. However, increased fertility also implies greater demand for maternal health, early-childhood care, and education systems in the short term \cite{wodon2016investing}. These trade-offs are central to future planning and emphasise that demographic objectives must be carefully balanced with social and economic capacities.

In summary: 
\begin{enumerate}
    \item \textbf{Baseline trajectory (TFR $\approx 2.0$):} Maintains balanced age distribution but with high population levels. 
    \item \textbf{Declining fertility (TFR $\to 1.6$):} Accelerates ageing and increases dependency ratios. 
    \item \textbf{Policy boost (TFR $\to 2.2$):} Expands the labour force but increases resource pressures. 
    \item \textbf{Voluntary two-child policy:} Provides a middle path, balancing growth moderation with ageing risks. 
\end{enumerate}
Across all scenarios, the hybrid LSTM--PINN model confirms that age remains the dominant factor in population evolution, while policy-induced temporal changes shape the trajectory more gradually. The interpretability results reinforce that the model’s behaviour aligns with demographic theory, enabling policymakers to trust the projections when evaluating long-term strategies. Overall, the findings highlight the importance of designing fertility and family-planning policies that support sustainable demographic balance, ensuring both adequate workforce supply and manageable ageing trends. 
\newpage
\subsection{Population Pyramid Evolution under Baseline Scenario (2024–2054)}

\begin{figure}[h]
    \centering
    \begin{subfigure}[b]{0.45\textwidth}
        \centering
        \includegraphics[width=\textwidth]{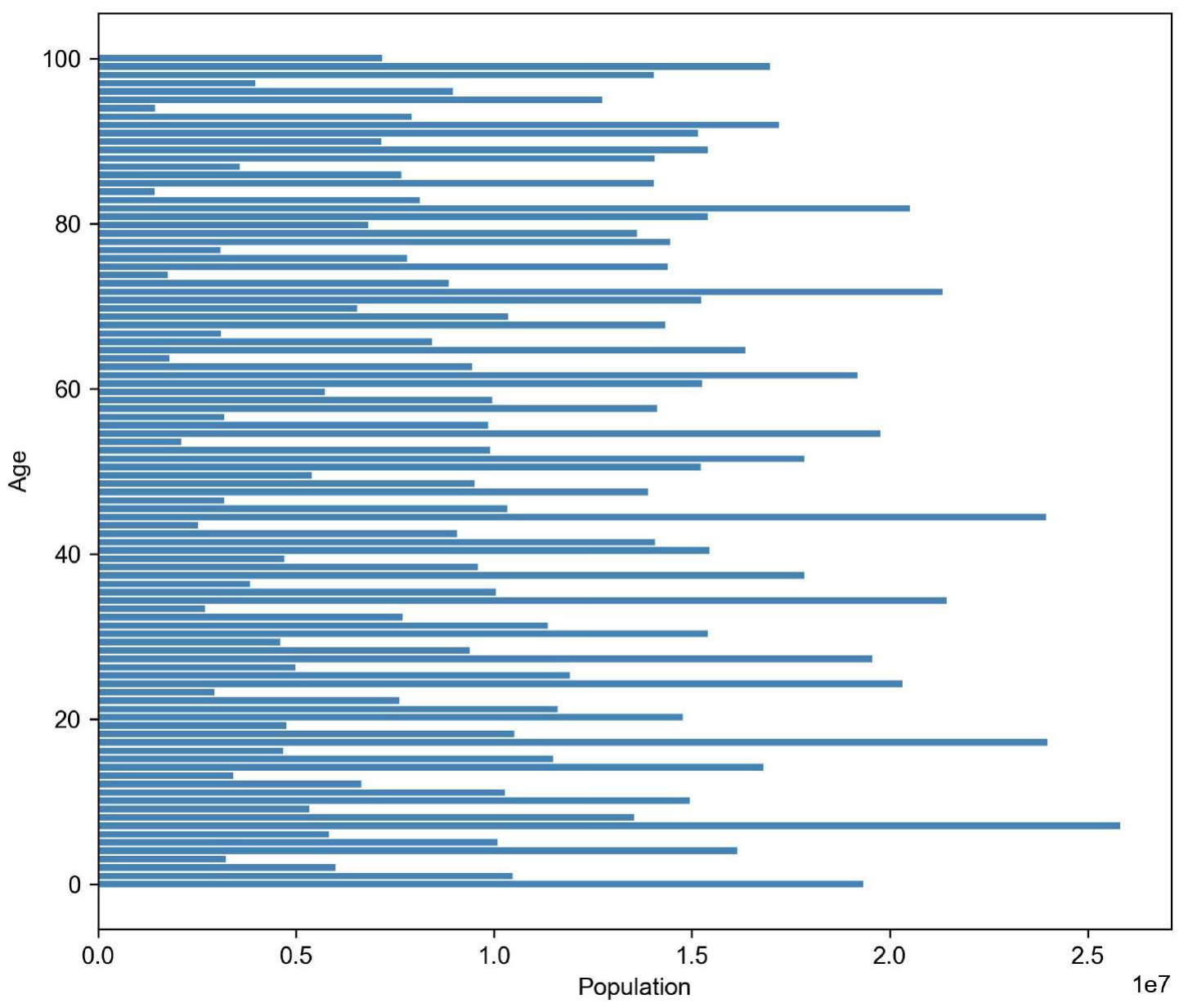}
        \caption{Population Pyramid (2024, Baseline)}
        \label{fig:pyr 1}
    \end{subfigure}
    \hfill
    \begin{subfigure}[b]{0.45\textwidth}
        \centering
        \includegraphics[width=\textwidth]{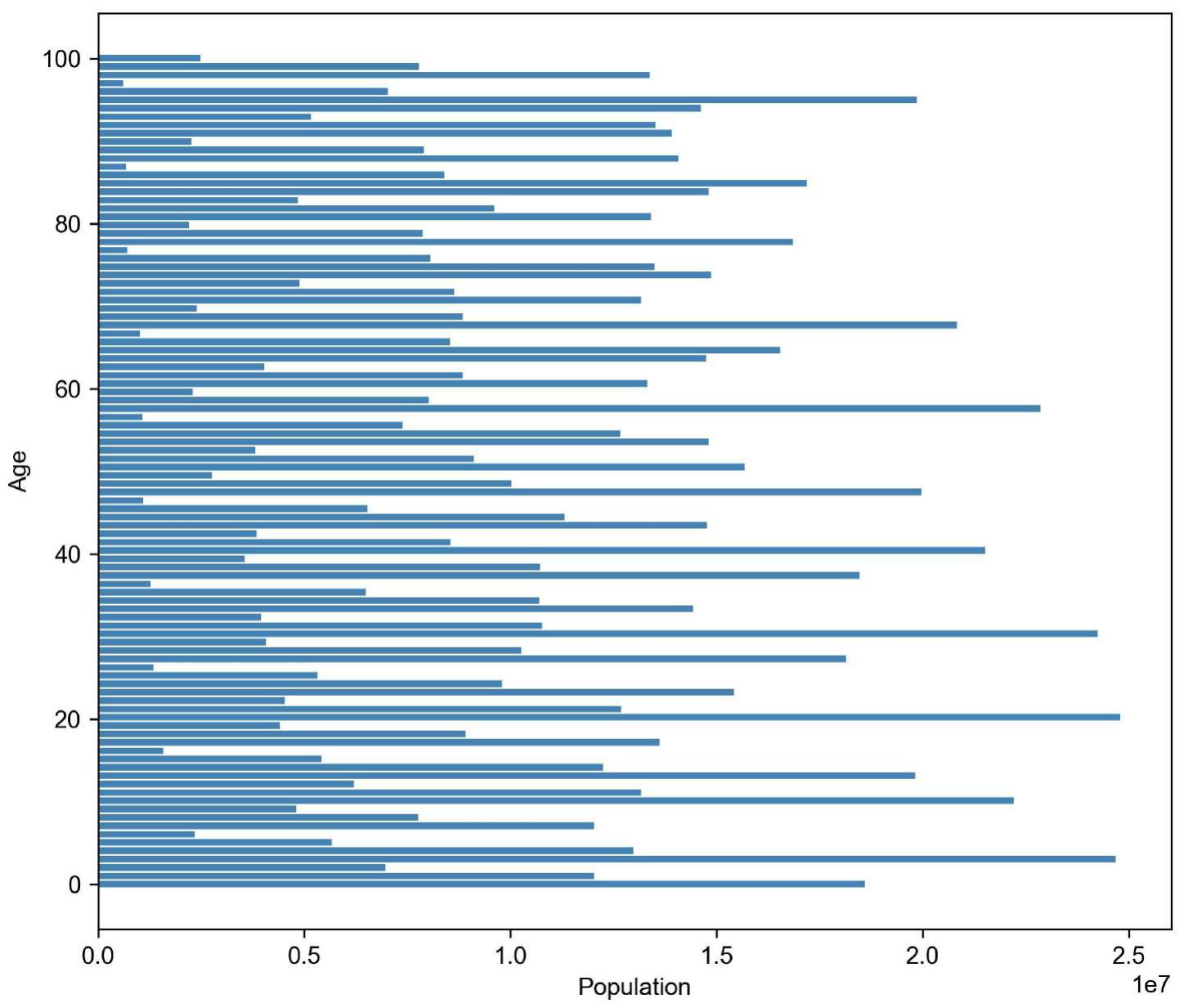}
        \caption{Population Pyramid (2039, Baseline)}
        \label{fig:pyr 2}
    \end{subfigure}
    \hfill
    \begin{subfigure}[b]{0.45\textwidth}
        \centering
        \includegraphics[width=\textwidth]{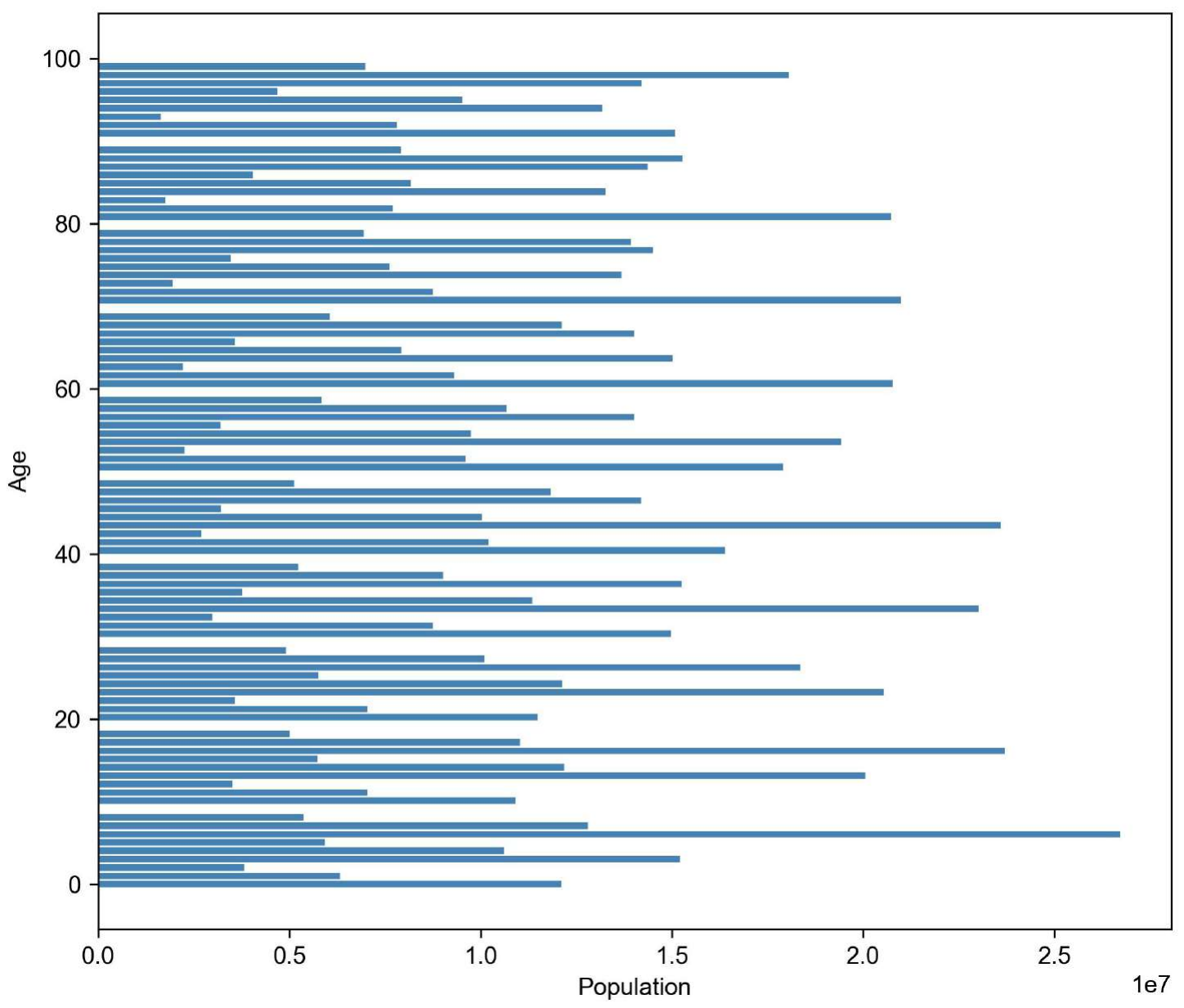}
        \caption{Population Pyramid (2054, Baseline)}
        \label{fig:pyr 3}
    \end{subfigure}
    \caption{Baseline population pyramids for India, illustrating the transition from a youthful structure to an ageing society.}
    \label{fig:pyr 2054}
\end{figure}

The population pyramids for 2024--2054 provide an intuitive visualisation of the demographic transition unfolding under the baseline fertility trajectory. As shown in Figure \ref{fig:pyr 1}, the pyramid in 2024 features a broad youth base and a pronounced working-age bulge, reflecting India’s youthful starting structure and strong demographic momentum. This wide base supplies the large cohorts that shape all subsequent projections in the baseline scenario.

By the mid-period (Figure \ref{fig:pyr 2}), the youth base begins to narrow while middle-age cohorts remain substantial, indicating the early stages of population ageing. This shift highlights mid-term policy challenges, particularly regarding labour-force sustainability, employment generation, and skill development.

The projection for 2054 (Figure \ref{fig:pyr 3}) shows a markedly contracted youth base alongside a visibly expanding elderly population, signalling advanced demographic ageing. This long-term pattern underscores the rising pressure on pension systems, healthcare infrastructure, and elder-care services. Together, the pyramids reveal a clear and gradual transition from a youthful population to an ageing society, consistent with the trends captured by both the PINN and LSTM--PINN models in earlier subsections.

\newpage
\section{Conclusion and future work}

This study introduced a hybrid LSTM--PINN framework for forecasting India’s demographic evolution from 2024 to 2054, combining the mechanistic grounding of age-structured population dynamics with the temporal learning capacity of neural networks. By incorporating India-specific fertility, mortality, and migration inputs, together with policy-aware fertility functions, the model generated coherent projections across multiple demographic scenarios. The \textit{results} suggest that a baseline fertility trajectory (TFR $\approx 2.0$) supports gradual stabilisation, while continued fertility decline (TFR $\to 1.6$) accelerates ageing and constrains the future labour force. Alternatively, modest fertility-enhancing interventions (TFR $\to 2.2$) alleviate but do not fully reverse long-term ageing trends. These outcomes reinforce the need for balanced policies that support both demographic sustainability and socioeconomic planning.

The model's age-structured projections, population pyramids, and consistently convergent loss behaviour demonstrate the effectiveness of the hybrid LSTM--PINN architecture in capturing nonlinear demographic dependencies. The fusion of PDE-based interpretability with LSTM-driven temporal refinement yields a flexible forecasting framework that respects demographic theory while accommodating policy shocks and long-range temporal patterns. The findings highlight key policy considerations for India: a narrowing youth base, expanding elderly population, and regional variations in fertility response demand targeted interventions in healthcare, labour markets, family planning, and social protection systems.

\textbf{Limitations and Future Work:}  
While the proposed framework advances demographic forecasting, several limitations offer directions for further improvement. First, migration was modelled as an exogenous adjustment; integrating an endogenous, age-specific migration module could enhance realism, particularly for states with substantial internal mobility. Second, the model assumes nationally aggregated fertility and mortality schedules; future work could extend the framework to a state-level or district-level hierarchical architecture to capture India’s substantial demographic heterogeneity. Third, although the LSTM component learns temporal patterns effectively, emerging architectures such as Transformers or Neural ODEs may further improve long-horizon temporal reasoning. Additionally, uncertainty quantification, such as Bayesian PINNs or ensemble-based LSTM approaches, would strengthen the robustness of policy-sensitive projections. Finally, integrating behavioural responses (education, employment, gender norms) or socioeconomic feedback loops could make the model an even more comprehensive tool for demographic and development policy analysis.

Overall, this study establishes a novel LSTM--PINN framework that enhances both the accuracy and interpretability of population projections. While developed for India, the approach is generalizable to other countries undergoing demographic transition, offering a valuable methodological contribution to computational demography and long-term policy planning.

 \bibliographystyle{elsarticle-num} 
 \bibliography{myref}

\begin{thebibliography}{10}
\expandafter\ifx\csname url\endcsname\relax
  \def\url#1{\texttt{#1}}\fi
\expandafter\ifx\csname urlprefix\endcsname\relax\def\urlprefix{URL }\fi
\expandafter\ifx\csname href\endcsname\relax
  \def\href#1#2{#2} \def\path#1{#1}\fi

\bibitem{asao2024japan}
K.~Asao, D.~Smirnov, M.~T. Xu, Japan’s Fertility: More Children Please, International Monetary Fund, 2024.

\bibitem{chen2023changing}
W.~Chen, Changing fertility patterns in china, Chinese Journal of Sociology 9~(4) (2023) 497--521.

\bibitem{bongaarts2022fertility}
J.~Bongaarts, D.~Hodgson, Fertility transition in the developing world, Springer Nature, 2022.

\bibitem{bernardi2025heterogeneously}
E.~Bernardi, T.~Lorenzi, M.~Sensi, A.~Tosin, Heterogeneously structured compartmental models of epidemiological systems: From individual-level processes to population-scale dynamics, Studies in Applied Mathematics 155~(2) (2025) e70091.

\bibitem{diekmann2023systematic}
O.~Diekmann, H.~Inaba, A systematic procedure for incorporating separable static heterogeneity into compartmental epidemic models, Journal of Mathematical Biology 86~(2) (2023) 29.

\bibitem{zhao2023comparative}
Y.~Zhao, S.~W. Wong, A comparative study of compartmental models for covid-19 transmission in ontario, canada, Scientific Reports 13~(1) (2023) 15050.

\bibitem{losanova2023boundary}
F.~M. Losanova, R.~O. Kenetova, Boundary value problem for the loaded mckendrick von foerster equation of fractional order, Adyghe International Scientific Journal 23~(4) (2023) 28--33.

\bibitem{chen2025data}
B.~Chen, R.~Guo, Q.~Zhang, Y.~Zhao, X.~Wang, Z.~Zhu, A data-driven crowd simulation framework integrating physics-informed machine learning with navigation potential fields, IEEE Transactions on Computational Social Systems (2025).

\bibitem{halder2023numerical}
J.~Halder, S.~K. Tumuluri, Numerical solution to a nonlinear mckendrick-von foerster equation with diffusion, Numerical Algorithms 92~(2) (2023) 1007--1039.

\bibitem{halder2024higher}
J.~Halder, S.~K. Tumuluri, A higher order numerical scheme to a nonlinear mckendrick--von foerster equation with singular mortality, Applied Numerical Mathematics 202 (2024) 21--41.

\bibitem{raissi2019physics}
M.~Raissi, P.~Perdikaris, G.~E. Karniadakis, Physics-informed neural networks: A deep learning framework for solving forward and inverse problems involving nonlinear partial differential equations, Journal of Computational physics 378 (2019) 686--707.

\bibitem{karniadakis2021physics}
G.~E. Karniadakis, I.~G. Kevrekidis, L.~Lu, P.~Perdikaris, S.~Wang, L.~Yang, Physics-informed machine learning, Nature Reviews Physics 3~(6) (2021) 422--440.

\bibitem{tao2025analytical}
Y.~Tao, C.~Xiao, H.~Wang, Analytical solution to emi response of high-frequency field-line coupling with lumped-load based on asymptotic theory, in: 2025 IEEE 8th International Electrical and Energy Conference (CIEEC), IEEE, 2025, pp. 2174--2179.

\bibitem{hochreiter1997long}
S.~Hochreiter, J.~Schmidhuber, Long short-term memory, Neural computation 9~(8) (1997) 1735--1780.

\bibitem{gers2002learning}
F.~A. Gers, N.~N. Schraudolph, J.~Schmidhuber, Learning precise timing with lstm recurrent networks, Journal of machine learning research 3~(Aug) (2002) 115--143.

\bibitem{cho2022lstm}
G.~Cho, D.~Zhu, J.~J. Campbell, M.~Wang, An lstm-pinn hybrid method to estimate lithium-ion battery pack temperature, Ieee Access 10 (2022) 100594--100604.

\bibitem{kim2025performance}
S.~Kim, D.~Lee, S.~Lee, Performance improvement of seismic response prediction using the lstm-pinn hybrid method, Biomimetics 10~(8) (2025) 490.

\bibitem{wang5101380hybrid}
M.~Wang, H.~Chen, E.~Zhu, S.~Shi, P.~Huang, X.~Fang, X.~Wei, H.~Nie, A hybrid multiaxial fatigue life prediction method based on lstm-pinn, Available at SSRN 5101380 (2024).

\bibitem{jiang2025neighbour}
C.~Jiang, Q.~Bai, M.~Hardie, A neighbour-aware lstm-pinn model for physically consistent prediction of soil moisture and water retention curves, IEEE (2025).

\bibitem{zhang2025temperature}
Y.~Zhang, Z.~Zhou, Z.~Zhang, Z.~Liang, H.~Duan, Y.~Li, Temperature drift compensation method for fiber-optic gyroscope based on lstm-pinn, IEEE Photonics Technology Letters (2025).

\bibitem{gandotra1998fertility}
M.~Gandotra, R.~D. Retherford, A.~Pandey, N.~Y. Luther, V.~K. Mishra, Fertility in india, International Institute for Population Sciences (1998).

\bibitem{mahapatra2020migration}
B.~Mahapatra, N.~Saggurti, R.~Mishra, M.~Walia, S.~Mukherjee, Migration and family planning in the state with highest total fertility rate in india, BMC public health 20~(1) (2020) 1826.

\bibitem{dharmalingam2014determinants}
A.~Dharmalingam, S.~Rajan, S.~P. Morgan, The determinants of low fertility in india, Demography 51~(4) (2014) 1451--1475.

\bibitem{suriyakala2016factors}
V.~Suriyakala, M.~Deepika, J.~Amalendu, G.~Deepa, Factors affecting infant mortality rate in india: an analysis of indian states, in: The International Symposium on Intelligent Systems Technologies and Applications, Springer, 2016, pp. 707--719.

\bibitem{subaiya2011demographics}
L.~Subaiya, D.~W. Bansod, Demographics of population ageing in India, Institute for Social and Economic Change, 2011.

\bibitem{james2011india}
K.~S. James, India’s demographic change: opportunities and challenges, Science 333~(6042) (2011) 576--580.

\bibitem{rouyer1987political}
A.~R. Rouyer, Political capacity and the decline of fertility in india, American Political Science Review 81~(2) (1987) 453--470.

\bibitem{aspalter2002population}
C.~Aspalter, Population policy in india, International journal of sociology and social policy 22~(11/12) (2002) 48--72.

\bibitem{diamond2010population}
N.~Diamond-Smith, M.~Potts, Are the population policies of india and china responsible for the fertility decline?, International Journal of Environmental Studies 67~(3) (2010) 291--301.

\bibitem{jindal2018mid}
U.~N. Jindal, Mid-life fertility: Challenges \& policy planning, Indian Journal of Medical Research 148~(Suppl 1) (2018) S15--S26.

\bibitem{tao2025lstm}
Z.~Tao, An lstm-pinn hybrid method to the specific problem of population forecasting, arXiv preprint arXiv:2505.01819 (2025).

\bibitem{chakravorty2021family}
S.~Chakravorty, S.~Goli, K.~S. James, Family demography in india: Emerging patterns and its challenges, Sage Open 11~(2) (2021) 21582440211008178.

\bibitem{keyfitz1997mckendrick}
B.~L. Keyfitz, N.~Keyfitz, The mckendrick partial differential equation and its uses in epidemiology and population study, Mathematical and Computer Modelling 26~(6) (1997) 1--9.

\bibitem{mico2006age}
J.~C. Mico, D.~Soler, A.~Caselles, Age-structured human population dynamics, Journal of Mathematical Sociology 30~(1) (2006) 1--31.

\bibitem{dilao2006weak}
R.~Dilao, A.~Lakmeche, On the weak solutions of the mckendrick equation: Existence of demography cycles, Mathematical Modelling of Natural Phenomena 1~(1) (2006) 1--30.

\bibitem{bala2024effective}
I.~Bala, T.-L. Kelly, R.~Lim, M.~H. Gillam, L.~Mitchell, An effective approach for multiclass classification of adverse events using machine learning, Journal of Computational and Cognitive Engineering 3~(3) (2024) 226--239.

\bibitem{lawal2025modeling}
Z.~K. Lawal, H.~Yassin, D.~T.~C. Lai, A.~C. Idris, Modeling the complex spatio-temporal dynamics of ocean wave parameters: A hybrid pinn-lstm approach for accurate wave forecasting, Measurement 252 (2025) 117383.

\bibitem{kiyani2025optimizing}
E.~Kiyani, K.~Shukla, J.~F. Urb{\'a}n, J.~Darbon, G.~E. Karniadakis, Optimizing the optimizer for physics-informed neural networks and kolmogorov-arnold networks, Computer Methods in Applied Mechanics and Engineering 446 (2025) 118308.

\bibitem{chandrasekhar2006demographic}
C.~Chandrasekhar, J.~Ghosh, A.~Roychowdhury, The'demographic dividend'and young india's economic future, Economic and Political Weekly (2006) 5055--5064.

\bibitem{jain2025population}
N.~Jain, S.~Goli, A.~Jana, Population age structural transition, demographic dividend and economic growth in india, Humanities and Social Sciences Communications 12~(1) (2025) 1--13.

\bibitem{chatterjee2001planning}
N.~Chatterjee, N.~E. Riley, Planning an indian modernity: The gendered politics of fertility control, Signs: Journal of women in culture and society 26~(3) (2001) 811--845.

\bibitem{mandelbaum1974human}
D.~G. Mandelbaum, Human fertility in India: Social components and policy perspectives, Univ of California Press, 1974.

\bibitem{brakman2025demography}
S.~Brakman, T.~Kohl, C.~van Marrewijk, Demography and income in the 21st century: a long-run perspective, Cambridge Journal of Regions, Economy and Society 18~(1) (2025) 25--40.

\bibitem{world1978world}
W.~B. I. E. D. D.~D. Group, W.~B. I. E. D. D.~D. Group, World development indicators, World Bank, 1978.

\bibitem{wang2021health}
M.~Wang, Health indicators on adolescents reveal disparity and inequality on regional and national levels, BMC Public Health 21~(1) (2021) 1006.

\bibitem{rao2021premature}
C.~Rao, A.~Gupta, M.~Gupta, A.~K. Yadav, Premature adult mortality in india: what is the size of the matter?, BMJ Global Health 6~(6) (2021) e004451.

\bibitem{maiti2023socioeconomic}
S.~Maiti, S.~Akhtar, A.~K. Upadhyay, S.~K. Mohanty, Socioeconomic inequality in awareness, treatment and control of diabetes among adults in india: Evidence from national family health survey of india (nfhs), 2019--2021, Scientific reports 13~(1) (2023) 2971.

\bibitem{jayaraman2013demographic}
A.~Jayaraman, A demographic overview, State of the urban youth, India (2013) 8--15.

\bibitem{bhalla2011labour}
S.~Bhalla, R.~Kaur, Labour force participation of women in india: some facts, some queries, Labour Force Participation of Women (2011).

\bibitem{kalyani2015unorganised}
M.~Kalyani, Unorganised workers: A core strength of indian labour force: An analysis, International Journal of Research 44 (2015).

\bibitem{desa2022world}
U.~Desa, World population prospects 2022, NYUN Dep Econ Soc Aff Popul Div Httpesa Un Orgunpdwpp (2022).

\bibitem{escapil2023h}
P.~Escapil-Inchausp{\'e}, G.~A. Ruz, h-analysis and data-parallel physics-informed neural networks, Scientific Reports 13~(1) (2023) 17562.

\bibitem{sharma2022accelerated}
R.~Sharma, V.~Shankar, Accelerated training of physics-informed neural networks (pinns) using meshless discretizations, Advances in neural information processing systems 35 (2022) 1034--1046.

\bibitem{ribeiro2016should}
M.~T. Ribeiro, S.~Singh, C.~Guestrin, " why should i trust you?" explaining the predictions of any classifier, in: Proceedings of the 22nd ACM SIGKDD international conference on knowledge discovery and data mining, 2016, pp. 1135--1144.

\bibitem{bloom2001economic}
D.~E. Bloom, D.~Canning, J.~Sevilla, Economic growth and the demographic transition, National Bureau of Economic Research (2001).

\bibitem{bloom2003demographic}
D.~Bloom, D.~Canning, J.~Sevilla, The demographic dividend: A new perspective on the economic consequences of population change, Rand Corporation, 2003.

\bibitem{wodon2016investing}
Q.~Wodon, Investing in early childhood development: essential interventions, family contexts, and broader policies, Journal of Human Development and Capabilities 17~(4) (2016) 465--476.

\end{thebibliography}

\end{document}